%% file: 1-main.tex
\documentclass[10pt,twocolumn,letterpaper]{article}

\usepackage{cvpr}
\usepackage{graphicx}
\usepackage{grffile}
\usepackage{amsmath,amssymb,mathtools,amsthm}
\usepackage{booktabs}
\usepackage{lipsum}
\usepackage[dvipsnames]{xcolor}
\usepackage{animate}

\usepackage{xspace}

\usepackage{algpseudocode}
\usepackage{algorithm}
\usepackage{enumitem}
\usepackage{footnote}
\definecolor{linkc}{rgb}{0, 0.44, 0.74}
\definecolor{eqc}{rgb}{1, 0, 0}
\definecolor{mygray}{gray}{0.92}
\definecolor{baselinecolor}{gray}{.9}
\newcommand{\baseline}[1]{\cellcolor{baselinecolor}{#1}}
\usepackage{listings}
\usepackage{amsthm}
\usepackage{tabulary}
\usepackage{colortbl}
\usepackage{enumitem}
\usepackage{braket}
\usepackage{array}
\usepackage{comment}
\usepackage{caption}
\usepackage{subcaption}

\usepackage{float}
\usepackage{enumitem}
\usepackage{booktabs}
\usepackage{algorithm}
\usepackage{bbding}
\usepackage{listings}
\usepackage{tabu}
\usepackage{float}

\usepackage{microtype}
\usepackage{graphicx}
\usepackage{booktabs} 
\usepackage{microtype}      % microtypography
\usepackage{xcolor}         % colors
\usepackage{color}
\usepackage{tabulary,multirow,overpic,xcolor}

\newcommand{\adamae}{\textit{Ada}MAE\xspace}

\usepackage{pifont} 
\newcommand{\cmark}{\ding{51}}%
\newcommand{\xmark}{\ding{55}}%

\def\x{$\times$}

\newcolumntype{x}[1]{>{\centering\arraybackslash}p{#1pt}}
\newcolumntype{y}[1]{>{\raggedright\arraybackslash}p{#1pt}}
\newcolumntype{z}[1]{>{\raggedleft\arraybackslash}p{#1pt}}

\newlength\savewidth\newcommand\shline{\noalign{\global\savewidth\arrayrulewidth
		\global\arrayrulewidth 1pt}\hline\noalign{\global\arrayrulewidth\savewidth}}
\newcommand{\tablestyle}[2]{\setlength{\tabcolsep}{#1}\renewcommand{\arraystretch}{#2}\centering\footnotesize}

\definecolor{linkcol}{RGB}{233, 4, 141}

\definecolor{xycolor}{RGB}{60, 120, 216}
\definecolor{xycolor}{HTML}{0071bc}

\definecolor{wcolor}{RGB}{103, 78, 167}

\definecolor{dcolor}{RGB}{166, 77,21}

\definecolor{gcolor}{RGB}{204, 102, 153}

\definecolor{tcolor}{RGB}{34,139,34}

\definecolor{iterc}{RGB}{91,196,159}

\definecolor{epochc}{RGB}{96,172,252}

\definecolor{eicolor}{RGB}{153, 51, 102}

\usepackage[pagebackref,breaklinks,colorlinks]{hyperref}

\usepackage{listings}
\usepackage{lstautogobble}  %
\usepackage{color}          %
\usepackage{zi4}            %
\definecolor{bluekeywords}{rgb}{0.13, 0.13, 1}
\definecolor{greencomments}{rgb}{0, 0.5, 0}
\definecolor{redstrings}{rgb}{0.9, 0, 0}
\definecolor{graynumbers}{rgb}{0.5, 0.5, 0.5}

\usepackage[capitalize]{cleveref}
\crefname{section}{Sec.}{Secs.}
\Crefname{section}{Section}{Sections}
\Crefname{table}{Table}{Tables}
\crefname{table}{Tab.}{Tabs.}

\begin{document}
%%%%%%%%% TITLE - PLEASE UPDATE
\title{\adamae: Adaptive Masking for Efficient Spatiotemporal \\Learning with Masked Autoencoders}
\author{Wele Gedara Chaminda Bandara$^1$, Naman Patel$^2$, Ali Gholami$^2$, \\
Mehdi Nikkhah$^2$, Motilal Agrawal$^2$, and Vishal M. Patel$^1$ \vspace{2mm}\\
$^1$Johns Hopkins University, Baltimore, USA \hspace{10mm} $^2$Zippin, California, USA\\
{\tt\small wbandar1@jhu.edu, \{naman, gholami, mehdi, moti\}@getzippin.com, vpatel36@jhu.edu}\\
{
\href{https://github.com/wgcban/adamae.git}{\texttt{github.com/wgcban/adamae}}}
}
\maketitle

%%%%%%%%%%%%%%%%%%%%%%
%%%%%% Abstract %%%%%%
%%%%%%%%%%%%%%%%%%%%%%
\begin{abstract}
Masked Autoencoders (MAEs) learn generalizable representations for image, text, audio, video, etc., by reconstructing masked input data from tokens of the visible data. Current MAE approaches for videos rely on random patch, tube, or frame based masking strategies to select these tokens. This paper proposes \adamae, an adaptive masking strategy for MAEs that is end-to-end trainable. Our adaptive masking strategy samples visible tokens based on the semantic context using an auxiliary sampling network. This network estimates a categorical distribution over spacetime-patch tokens. The tokens that  increase the expected reconstruction error are rewarded and selected as visible tokens, motivated by the policy gradient algorithm in reinforcement learning. We show that \adamae samples more tokens from the high spatiotemporal information regions, thereby allowing us to mask 95\% of tokens, resulting in lower memory requirements and faster pre-training. We conduct ablation studies on the Something-Something v2 (SSv2) dataset to demonstrate the efficacy of our adaptive sampling approach and report state-of-the-art results of 70.0\% and 81.7\% in top-1 accuracy on SSv2 and Kinetics-400 action classification datasets with a ViT-Base backbone and 800 pre-training epochs.
\end{abstract}

%%%%%%%%%%%%%%%%%%%%
%%% Introduction %%%
%%%%%%%%%%%%%%%%%%%%
\section{Introduction}
\label{sec:intro}
%%% introduction figure %%%
\begin{figure}[tb]
    \centering
    \includegraphics[width=\linewidth]{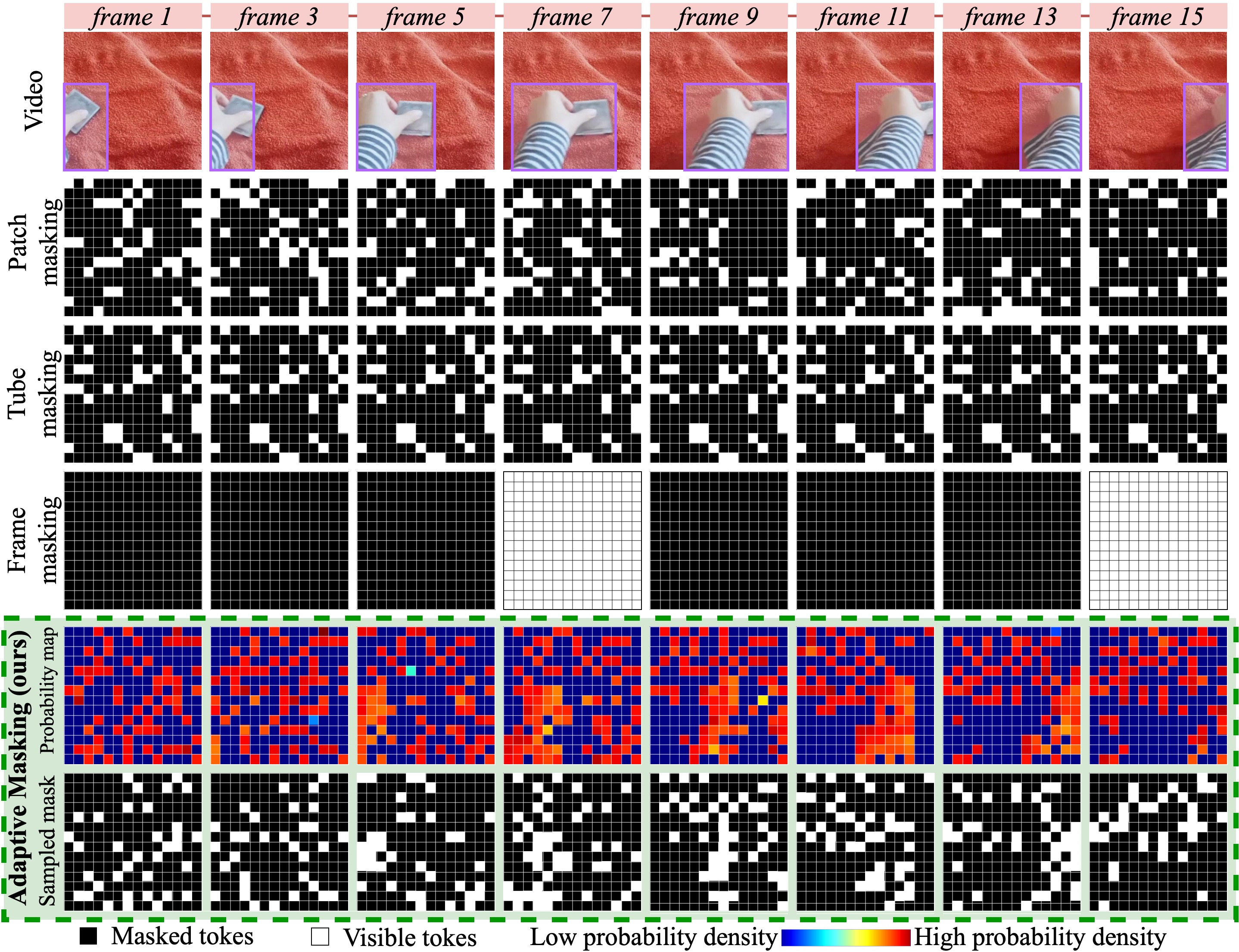}
    \caption{\textbf{Comparison of our \textit{adaptive masking} with existing random \textit{patch}~\cite{feichtenhofer_masked_2022}, \textit{tube}~\cite{tong_videomae_2022, wang_bevt_2022}, and \textit{frame}~\cite{lee_stochastic_2020, oord_representation_2019,shelhamer_loss_2017} masking for masking ratio of 80\%.} Our adaptive masking approach selects more tokens from the regions with \textcolor{violet}{\textit{ high spatiotemporal information}} while a small number of tokens from the background.
    %as the visible tokens to the MAE.
    }
    \label{fig:mask_type_comp}
\end{figure}
%%%%%%%%%%%%%%%%%%%%%%%%%%%
\par Self-supervised learning (SSL) aims to learn transferable representations from a large collection of unlabeled data for downstream applications (e.g., classification and detection). SSL is conducted in a two-stage framework~\cite{jaiswal_survey_2021}, consisting of pre-training on an unlabeled dataset, and fine-tuning on a downstream task.
Pre-training has shown to improve performance~\cite{feichtenhofer_masked_2022},  convergence speed~\cite{feichtenhofer_masked_2022}, and robustness~\cite{hendrycks_using_2019}, and reduce model overfitting~\cite{feichtenhofer_masked_2022, girdhar_omnimae_2022} on downstream tasks. 

\par Recently, masked autoencoders (MAEs)~\cite{tong_videomae_2022, gupta_maskvit_2022, xie_masked_2022, feichtenhofer_masked_2022, he_masked_2021, huang_green_2022, chen_efficient_2022-1} and contrastive learning~\cite{tao_siamese_2022, qian_spatiotemporal_2021, jaiswal_survey_2021} approaches are mainly used for SSL. In \textit{MAEs}, the input (image or video) is patchified and converted into a set of tokens. A small percentage (\eg, 5-10\%) of these tokens, namely visible tokens, are passed through a Vision Transformer (ViT) ~\cite{dosovitskiy_image_2021}. The resulting token embeddings are then concatenated with a learnable representation for masked tokens, and are fed into a shallow decoder transformer to reconstruct masked patches. On the other hand, \textit{contrastive learning} takes two augmented views of the same input and pulls them together in the embedding space, while embeddings of different inputs are pushed away~\cite{qian_spatiotemporal_2021}. MAEs have recently gained more attention over contrastive learning methods due to the inherent use of a high masking ratio, which enables simple and memory-efficient training.
% The high masking ratio employed by MAEs enables a simple and memory-efficient training leading to their recent gain in attention and wide adoption
% Despite the initial success of contrastive learning methods, MAEs have recently gained more attention and have been widely adopted due to their simple and memory-efficient training enabled by the high masking ratio.
%More importantly, MAEs makes the pre-training ViT more efficient and faster specially when pre-training it on videos as the number of tokens for a given video is much higher than the same size image, making it difficult to utilize with methods that takes all tokens as the input as it drastically slow down the training process. 

\par Mask sampling techniques are critical to the success of MAEs~\cite{feichtenhofer_masked_2022, wei_masked_2021}. Previous studies have investigated different sampling techniques that include random  ``patch''~\cite{feichtenhofer_masked_2022}, ``tube''~\cite{wei_masked_2021}, and ``frame''~\cite{wei_masked_2021} masking (see Fig. \ref{fig:mask_type_comp}). Random patch sampling has shown to work well compared to its counterparts in some cases~\cite{feichtenhofer_masked_2022}. However, since not all tokens have equal information, assuming a uniform probability distribution over all input tokens (for selection of visible tokens) is sub-optimal. In other words, with these random masking strategies, the visible tokens are sampled from redundant or low information regions instead of high information ones, hence resulting in inaccurate reconstructions.
This inhibits MAEs from learning meaningful representations, besides requiring a relatively larger number of training iterations compared to contrastive learning methods.

\par In this paper, we propose an \textit{adaptive sampling} approach that simultaneously optimizes an MAE and an adaptive token sampling network. Our approach selects patches based on their \textit{spatiotemporal information}. Unlike uniform random sampling, we first estimate the categorical distribution over all input tokens using an auxiliary network, and then sample visible tokens from that distribution. Since sampling is a non-differentiable operation, we propose an auxiliary loss for optimizing the adaptive token sampling network. Our solution is motivated by the REINFORCE algorithm~\cite{williams_simple_1992}, which comes under the family of policy gradient algorithms in Reinforcement Learning (RL)~\cite{kaelbling_reinforcement_1996}. 

We empirically show that our adaptive token sampling network leads to sampling more tokens from high spatiotemporal information regions compared to random masking techniques as shown in Fig.~\ref{fig:mask_type_comp}. This efficient token allocation also enables high masking ratios (i.e., 95\%) for pre-training MAEs. This ultimately reduces the GPU memory requirements and expedites pre-training while improving accuracy on downstream tasks.
In summary, our contributions are:

\begin{itemize}
    \item We propose \adamae, a novel, adaptive, and end-to-end trainable token sampling strategy for MAEs that takes into account the spatiotemporal properties of all input tokens to sample fewer but informative tokens.

    \item We empirically show that \adamae samples more tokens from high spatiotemporal information regions of the input, resulting in learning meaningful representations for downstream tasks.

    \item We demonstrate the efficiency of \adamae in terms of performance and GPU memory against random ``patch'', ``tube'', and ``frame'' sampling by conducting a thorough ablation study on the SSv2 dataset.

    \item We show that our \adamae outperforms state-of-the-art (SOTA) by \textbf{0.7\%} and \textbf{1.1\%} (in top-1) improvements on SSv2 and Kinetics-400, respectively.
    
\end{itemize}

%%%%%%%%%%%%%%%%%%%%%%%%%%%%%%%%%%%%%%%%%%%%%%%%%%%%%%%%%%%%%%%
%%%%%%%%%%%%%%%%%%%%%%%%% Related Work %%%%%%%%%%%%%%%%%%%%%%%%
%%%%%%%%%%%%%%%%%%%%%%%%%%%%%%%%%%%%%%%%%%%%%%%%%%%%%%%%%%%%%%%
\section{Related Work}
%%% Masked prediction ...
\subsection{Masked prediction}
\par Masked prediction has been widely successful in natural language understanding (e.g. Generative Pre-Training (GPT) ~\cite{radford_improving_nodate} and bidirectional encoder (BERT) ~\cite{devlin_bert_2019}). Many researchers have investigated applying masked prediction to representation learning for images and videos. Generative pre-training from pixels (iGPT) ~\cite{chen_generative_2020} is the first proof of concept in this direction, where   masked \textit{pixel} prediction is performed. However, this pixel-based method has a high pre-training computation cost and performs worse compared to ConvNets. With the introduction of the Vision Transformer (ViT) ~\cite{dosovitskiy_image_2021}, the focus shifted from pixels to patches. 
The concept of patches enables the ViTs to follow the masked language modeling task in BERT by predicting masked image patches (called iBERT) for visual pre-training. 

Although initial ViT experiments for representation learning have shown inferior performance compared to contrastive learning methods ~\cite{jing_understanding_2022, zhang_dual_2022, zhang_how_2022, qian_spatiotemporal_2021}, subsequent developments such as BERT pre-training including BEiT~\cite{bao_beit_2022-1}, BEVT~\cite{wang_bevt_2022}, and masked auto-encoders (MAEs) including MaskFeat~\cite{wei_masked_2021}, spatiotemporal-learner~\cite{he_masked_2021}, SimMIM~\cite{xie_masked_2022}, MFM~\cite{xie_masked_2022}, VideoMAE~\cite{tong_videomae_2022}, ConvMAE~\cite{gao_convmae_2022}, OmniMAE~\cite{girdhar_omnimae_2022}, MIMDET~\cite{fang_unleashing_2022}, paved the way for superior performance and lower pre-training times.

Since image patches do not have off-the-shelf tokens as words in languages, BEiT ~\cite{bao_beit_2022-1} and BEVT~\cite{wang_bevt_2022} adopt a two-stage pre-training strategy, where an image/video tokenizer is trained via a discrete variational auto-encoder (dVAE) ~\cite{ramesh_zero-shot_2021} followed by masked patch prediction using the pre-trained tokenizer~\cite{zhang_survey_2022}. This two-stage training  and the dependence on a pre-trained dVAE to generate originally continuous but intentionally discretized target visual tokens leaves room for improving the efficiency. On the other hand, MAEs~\cite{bao_beit_2022-1, he_masked_2021, xie_masked_2022, tong_videomae_2022, girdhar_omnimae_2022, fang_unleashing_2022} directly predict the masked patches from the visible tokens for pre-training. Moreover,  experiments suggest that a high masking ratio (75\% for images~\cite{bao_beit_2022-1, he_masked_2021} and 90\% for videos~\cite{xie_masked_2022, tong_videomae_2022})  leads to better fine-tuning performance. The encoder of MAEs only operates on the visible patches and the decoder is lightweight for faster pre-training.  This results in MAE pre-training which is  three times faster than BEiT~\cite{he_masked_2021, feichtenhofer_masked_2022, wettig_should_2022} and BEVT~\cite{wang_bevt_2022}.

%%% Mask sampling ...
\subsection{Mask sampling}
 Various ablation experiments of masked autoencoders on both images and videos have shown that the performance on downstream tasks depends on the masking strategy~\cite{xie_masked_2022, tong_videomae_2022, feichtenhofer_masked_2022}. Various random masking strategies have been proposed including  \textit{grid}, \textit{block}, and \textit{patch} for images ~\cite{he_masked_2021,xie_simmim_2022} and \textit{tube}, \textit{frame} and \textit{patch} for videos ~\cite{feichtenhofer_masked_2022, xie_masked_2022,lee_stochastic_2020, oord_representation_2019,shelhamer_loss_2017}.  Experiments have shown that a mask sampling technique that works well on one dataset might not be ideal for another dataset. For instance, Video-MAE~\cite{tong_videomae_2022} achieves the best action classification performance on the SSv2~\cite{goyal_something_2017} dataset with random tube masking, whereas SpatiotemporalMAE \cite{feichtenhofer_masked_2022} achieves its best performance on  Kinetics-400~\cite{kay_kinetics_2017} with random patch masking. This could be due to the diversity of underlying scenes in the dataset, different acquisition conditions (\eg, frame rate), and high/low spatiotemporal information regions in videos. Even within a dataset, we observe significant diversity among the videos, where some videos have dynamic scenes, some have a stationary background and moving foreground (see Fig. \ref{fig:mask_type_comp}), and some videos have almost stationary scenes (no object is moving). Since the existing masking techniques sample tokens randomly from \textit{space} (tube), \textit{time} (frame), or \textit{spacetime} (patch) and do not depend on the underlying scene, it is possible that the decoder tries to reconstruct high spatiotemporal information regions when most of the visible tokens belong to low spatiotemporal information regions~\cite{hou_milan_2022}. Consequently, a large number of pre-training iterations are required to learn meaningful representations~\cite{tong_videomae_2022, feichtenhofer_masked_2022} for downstream tasks. Inspired by this phenomenon, we design an adaptive masking technique based on reinforcement learning that predicts the sampling distribution for a given video.

%%% Reinforcement learning ...
\subsection{Reinforcement learning}
\par Reinforcement learning (RL) ~\cite{kaelbling_reinforcement_1996} has become an increasingly popular research area due to its effectiveness in various tasks, including playing Atari games~\cite{mnih_playing_2013}, image captioning~\cite{xu_show_2016}, person re-identification~\cite{lan_deep_2018}, video summarization~\cite{song_category_2016, zhou_deep_2018}, etc. Policy gradient methods~\cite{williams_simple_1992} for RL focus on selecting optimal policies that maximize expected return by gradient descent. REINFORCE ~\cite{sutton_policy_1999} is commonly seen as the basis for policy gradient methods in RL. However, its application in MAEs has yet to be explored. Our sampling network learns a categorical distribution over all tokens to sample informative visible tokens for pre-training MAEs. We use REINFORCE policy gradient~\cite{sutton_policy_1999} method as an unbiased estimator for calculating gradients to update the sampling network parameters $\theta$.

\par When the probability density function (estimated by the sampling network $f_{\theta}$) is differentiable with respect to its parameters $\theta$, we need to sample an action and compute probabilities $p(\cdot)$ to implement policy gradient~\cite{sutton_policy_1999}:
\setlength{\belowdisplayskip}{0pt} \setlength{\belowdisplayshortskip}{0pt}
\setlength{\abovedisplayskip}{0pt} \setlength{\abovedisplayshortskip}{0pt}
\begin{equation}
    \Delta \theta = \alpha \cdot R \cdot \frac{\partial \log p (a)}{\partial \theta},
\end{equation}
where $R$ is the return, $p (a)$ is the probability of taking action $a$, and $\alpha$ is the learning rate. In our case, we  sample a mask (the \textit{action}) based on the probability distribution estimated by the sampling network, reconstruct masked tokens from the MAE (the \textit{environment}), and then utilize reconstruction error (the \textit{reward}) to compute the expected return.

%%%%%%%%%%%%%%%%%%%%%%%%
%%%%%%%%% METHOD %%%%%%%
%%%%%%%%%%%%%%%%%%%%%%%%

% Method figure
\begin{figure*}[htb]
    \centering
    \includegraphics[width=\linewidth]{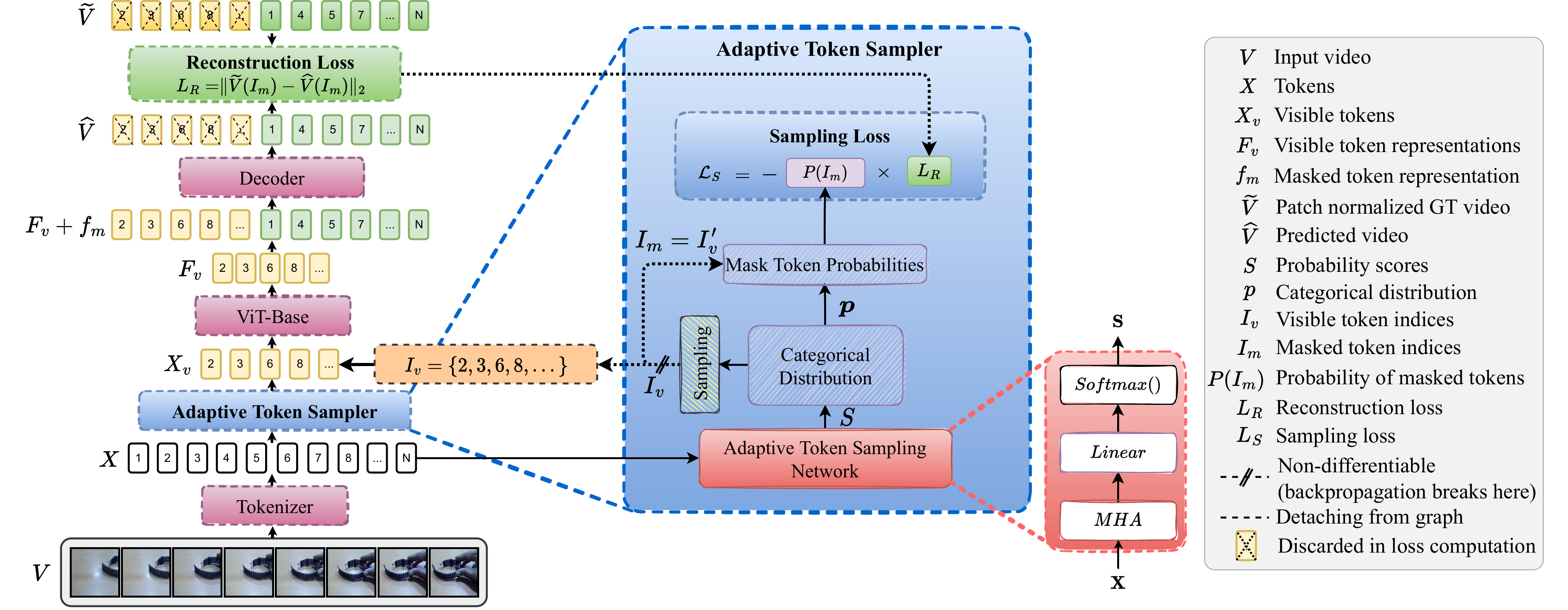}
    \caption{\textbf{\adamae} samples \textcolor{orange}{\textit{visible tokens}} based on the \textcolor{purple}{\textit{categorical distribution}} estimated by the \textcolor{blue}{\textit{sampling network}} and reconstructs the missing ones with a ViT encoder-decoder architecture. Since the sampling process is non-differentiable, we optimize it by \textbf{\textit{maximizing}} the \textcolor{blue}{\textit{expected reconstruction loss}}, which results in the sampling network predicting \textit{\textcolor{red}{high probability}} values for tokens that may lead to high reconstruction errors. Our adaptive sampling process samples more tokens from \textit{\textcolor{red}{high spatiotemporal information}} regions and fewer tokens from the \textit{\textcolor{blue}{low information}} or \textit{\textcolor{blue}{redundant}} regions.}
    \label{fig:method}
\end{figure*}

\section{Method}
% Architecture
\subsection{Architecture of \textbf{\adamae}}
Fig. \ref{fig:method} shows our \adamae architecture which consists of four main components, three of which (Tokenizer, Encoder, and Decoder) are standard to MAE along with our proposed Adaptive Token Sampler. 
%. Right dotted blue box shows our adaptive token sampling network which uses a shallow token probability estimation network to predict a categorical distribution of tokens based on reconstruction loss. 
\paragraph{Tokenizer:} Given an input video $\mathbf V$ of size $\textcolor{blue}{T} \times
\textcolor{red}{C} \times
\textcolor{teal}{H} \times \textcolor{teal}{W}$, where $\textcolor{blue}{T}$ denotes the number of temporal frames, $\textcolor{red}{C}$ denotes the input (RGB) channels, and $\textcolor{teal}{H} \times \textcolor{teal}{W}$ is the spatial resolution of a frame, we first pass it through a \texttt{Tokenizer} (i.e., 3D convolution layer with kernel $K$ of size $(\textcolor{blue}{t},
\textcolor{red}{C},
\textcolor{teal}{h}, \textcolor{teal}{w})$ and stride $S$ of size $(\textcolor{blue}{t},
\textcolor{teal}{h}, \textcolor{teal}{w})$, and $\textcolor{red}{d}$ output channels) to tokenize $\mathbf{V}$ into $N$ tokens of dimension $\textcolor{red}{d}$ denoted as $\mathbf X$, where $\textcolor{purple}{N} = \textcolor{blue}{\frac{T}{t}} \times \textcolor{teal}{\frac{H}{h}} \times \textcolor{teal}{\frac{W}{w}}$. Next, we inject the positional information into the tokens by utilizing the fixed 3D periodic positional encoding scheme introduced in \cite{vaswani_attention_2017}.

% Adaptive token sampling...
\paragraph{Adaptive Token Sampler:}
Given the tokens $\mathbf{X}$ resulting from the \texttt{Tokenizer}, we pass them through a light-weight Multi-Head Attention (MHA) network followed by a \texttt{Linear} layer and a \texttt{Softmax} activation to obtain the probability scores $\mathbf{P} \in \mathbb{R}^{ \textcolor{purple}{N}}$ for all tokens as follows:
\begin{align}
    \mathbf{Z} &= \texttt{MHA} (\mathbf{X}); \hspace{10pt}  \mathbf{Z} \in \mathbb{R}^{\textcolor{purple}{N} \times \textcolor{red}{d}},\\
    % logits &= \texttt{Linear} (\mathbf{Z}); \hspace{10pt} logits \in \mathbb{R}^{\textcolor{purple}{N} \times 1},\\
    % \mathbf{P} &= \texttt{Softmax} (logits); \hspace{10pt} \mathbf{P} \in \mathbb{R}^{\textcolor{purple}{N} \times 1}.
    \mathbf{P} &= \texttt{Softmax} (\texttt{Linear} (\mathbf{Z})); \hspace{10pt} \mathbf{P} \in \mathbb{R}^{\textcolor{purple}{N}}. 
\end{align}
We then assign an $N$-dimensional categorical distribution ~\cite{marriott__charles_dictionary_1990} over $\mathbf{P}$ ($\boldsymbol{p} \sim \text{Categorical}(N, \mathbf{P}))$ and draw (without replacement) a set of visible token indices  $\boldsymbol{I}_v$ (hence the set of masked token indices is given by $\boldsymbol{I}_m =  U-\boldsymbol{I}_v$, where $U$=$\{1, 2, 3, \cdots, \textcolor{purple}{N} \}$ is the set of all indices, i.e.,  $\boldsymbol{I}_m$ is the set complement of $\boldsymbol{I}_v$).
% as follows:
% \begin{align}
%     \boldsymbol{p} &\sim \text{Categorical}(N, \mathbf{P})
%     % \\
%     % \boldsymbol{I}_v &= \boldsymbol{p}.sample(N_v, \text{replacement}=False).
%     \label{eq: sampling}
% \end{align}
% Next, we sample $N_v$ visible tokens based on an $N$-dimensional categorical distribution $\boldsymbol{p}$ without replacement. 
The number of sampled visible tokens $N_v$ is computed based on a pre-defined masking ratio $\rho \in (0,1)$ and equals to $\textcolor{purple}{N} \times (1-\rho)$.

% We then select a set of visible tokens $\mathbf{X}_{v} \in \mathbb{R}^{\textcolor{purple}{N_s} \times \textcolor{red}{d}}$ by picking the tokens at visible indices $\boldsymbol{I}_v$ of $\mathbf{X}$.

% Encoder ...
\paragraph{Encoder:} Next, we generate latent representations $\mathbf{F}_{v}$ by passing sampled \textit{visible} tokens $\mathbf{X}_{v}$ through the \texttt{ViTEncoder}.

% Next, passing sampled \textit{visible} tokens $\mathbf{X}_{v}$ into the \texttt{ViTEncoder} generates their latent representation, $\mathbf{F}_{v}$.

% Decoder ...
\paragraph{Decoder:} The visible token representations $\mathbf{F}_{v}$ are concatenated with a fixed learnable representation $f_{m}$ for masked tokens. Next, positional information is added~\cite{qian_spatiotemporal_2021} for both representations, instead of shuffling them back to the original order. Finally, we pass this through a light-weight transformer decoder to obtain the predictions $\mathbf{\widehat{V}}$.

% Optimization ...
\subsection{Optimizing \textbf{\adamae}}
\paragraph{Masked reconstruction loss $\mathcal{L}_R$:} We utilize the mean squared error (MSE) loss between the predicted and \textit{patch normalized}\footnote{Each patch is normalized based on the patch mean and variance.} ground-truth RGB values~\cite{he_masked_2021} of the masked tokens to optimize the MAE (parameterzied by $\phi$) as follows:
\begin{equation}
    \begin{split}
     \mathcal{L}_{\text{R}}(\phi) & = \frac{1}{N- N_v} \sum_{i \in I_m} \lVert \mathbf{\widehat{V}}_i - \mathbf{\widetilde{V}}_i \rVert_2,
        \end{split}
\end{equation}
where $\widehat{\mathbf{V}}$ denotes the predicted tokens, $\mathbf{\widetilde{V}}$ denotes the local patch normalized ground-truth RGB values~\cite{tong_videomae_2022}.

% Adaptive sampling loss
\paragraph{Adaptive sampling loss $\mathcal{L}_S$:} We optimize the adaptive token sampling network (parameterized by $\theta$) using sampling loss  $\mathcal{L}_{\text{S}}$, which enables gradient update independent of the MAE (parameterized by $\phi$). The formulation of $\mathcal{L}_{\text{S}}$ is motivated by the REINFORCE algorithm in RL, where we consider the visible token sampling process as an \textit{action}, the MAE (ViT-backbone and decoders) as an \textit{environment}, and the masked reconstruction loss $\mathcal{L}_{R}$ as the \textit{return}. Following the expected reward maximization in the REINFORCE algorithm, we propose to optimize the sampling network by \textbf{maximizing} the  \textit{expected reconstruction error} $\mathbb{E}[\mathcal{L}_{R}]$. 

% figure for motivation behind the sampling loss
\begin{figure}[tb]
    \centering
    \includegraphics[width=\linewidth]{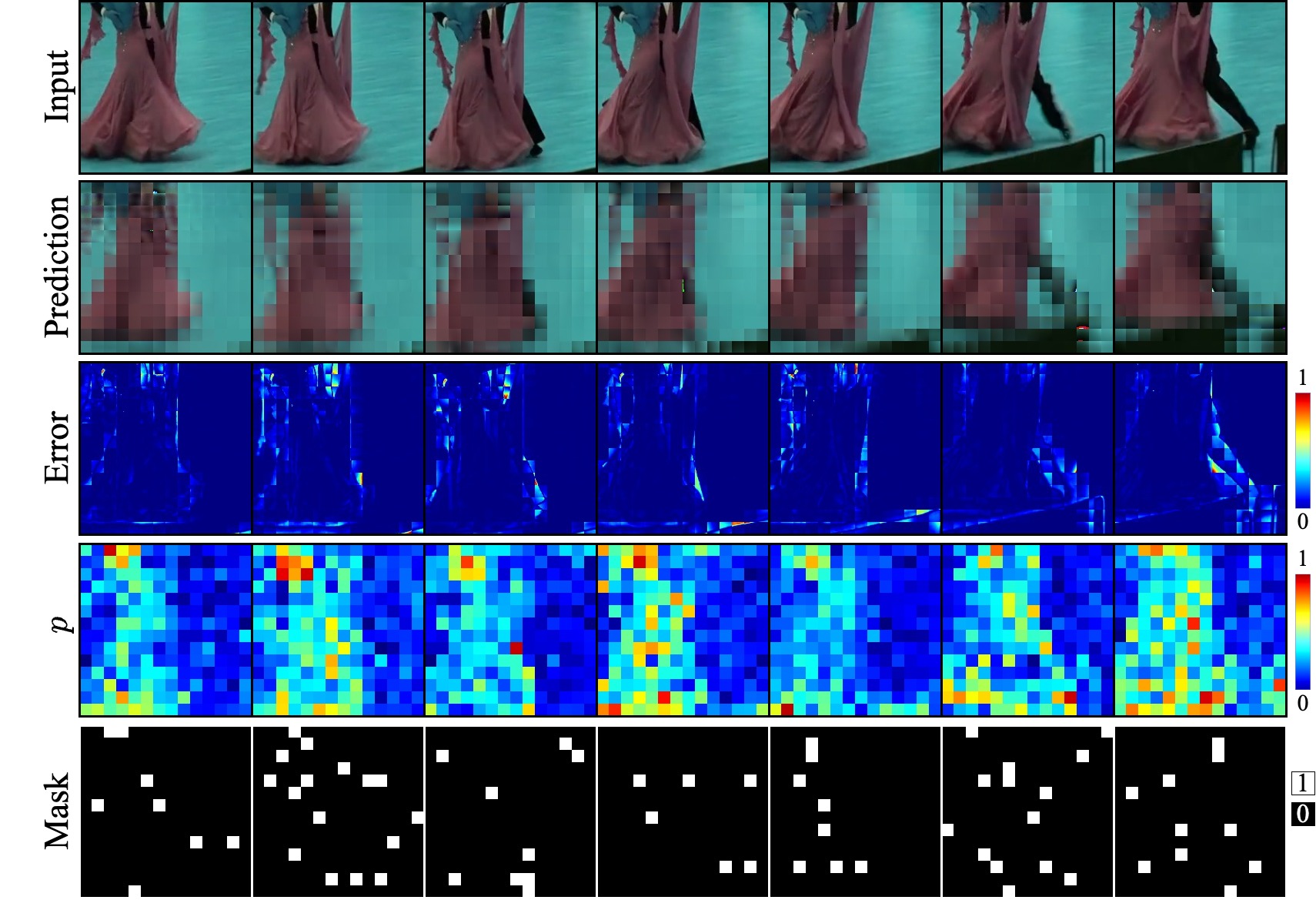}
    \caption{\textbf{Understanding the formulation of $L_S$.} We observe high $L_R$ in high activity/information regions (third row). Hence, we propose to  \textit{maximize expected reconstruction error} over masked tokens which ultimately results in the sampling network predicting \textcolor{red}{\textit{higher probability scores}} for \textcolor{red}{\textit{high activity}} regions and \textcolor{blue}{\textit{lower probability scores}} for \textcolor{blue}{\textit{low activity}} (i.e., background or redundant) regions. See Appendix for more visualizations.}
    \label{fig:loss_ls}
\end{figure}

\par To elucidate why maximizing $\mathbb{E}[\mathcal{L}_{R}]$ works for adaptive sampling, we use a visual example shown in Figure \ref{fig:loss_ls}. Given a video (\textit{first row}) with \textcolor{red}{\textit{high activity/information}} (i.e., the dancing couple) and \textcolor{blue}{\textit{low activity/information}} (i.e., the teal color background) regions, we observe a high reconstruction error (\textit{third row}) around the foreground. Since our objective is to sample more visible tokens from high activity regions and fewer tokens from the background, we optimize the sampling network by maximizing the expected reconstruction error over the masked tokens (denoted by $\mathbb{E}[\mathcal{L}_{R}]$). When optimized with the above rule, the adaptive token sampling network predicts \textcolor{red}{\textit{high probability scores}} for the tokens from the \textcolor{red}{\textit{high activity}} regions compared to the tokens from the background as shown in the fourth row. We observe that this adaptive token sampling approach for MAEs closely aligns with non-uniform sampling in compressed sensing, where more samples are assigned to the regions with high activity and fewer to those with low activity. Since our adaptive sampling allocates samples based on the level of spatiotemporal information, it requires fewer tokens to achieve the same reconstruction error compared to random sampling (which is not efficient in token allocation). This also enables using extremely high masking ratios (95\%) for MAEs which further reduces the computational burden, thereby accelerating the pre-training process. 

The objective function to optimize our adaptive token sampling network can be expressed as:
\begin{equation}
    \mathcal{L}_{S}(\theta)  = - \mathbb{E}_{\theta}[\mathcal{L}_{R}(\phi)] =  - \sum_{i \in \mathbf{I}_m} P_{\theta}^i \cdot \mathcal{L}_R^i (\phi),
\end{equation}
where $P_{\theta}^i$ is the probability of the mask token at index $i$ inferred from the adaptive token sampling network parameterized by $\theta$ and  $\mathcal{L}_R^i$ is the reconstruction error of the $i$-th masked token. Note that $\mathcal{L}_R(\phi)$ is the reconstruction error incurred by MAE with parameters $\phi$. Also, we explicitly prevent gradient updates from $\mathcal{L}_S(\theta)$ to propagate through the MAE (i.e., detach from the computational graph).

Furthermore, we take logarithm of the probabilities to avoid the precision errors caused by small probability values. Note that the negative sign is due to gradient descent optimization, whilst the rule above assumes gradient ascent.

%%%%%%%%%%%%%%%%%%%%%%%%%%%%%%%%%%
%%%%%%%%% EXPERIMENTS %%%%%%%%%%%%
%%%%%%%%%%%%%%%%%%%%%%%%%%%%%%%%%%
\section{Experimental Setup}
\paragraph{Datasets.} We evaluate \adamae on two datasets: Something-Something-v2 (SSv2) ~\cite{goyal_something_2017} and Kinetics-400~\cite{kay_kinetics_2017}. The SSv2 dataset is a relatively large-scale video dataset, consisting of approximately 170k videos for training, and 20k videos for validation, categorized into 174 action classes. Kinetics-400 is also a large-scale dataset which contains 400 action classes, $\sim$240k training videos, and $\sim$20k validation videos. We conduct ablation studies on the SSv2 dataset and report results on both SSv2 and Kinetics-400.

\paragraph{Dataset preprocessing.} We closely follow VideoMAE~\cite{tong_videomae_2022} for dataset preprocessing.  For both datasets, we sample 16 RGB frames, each with $224 \times 224$ pixels, from the raw video with a temporal stride of 4, where the starting frame location is randomly selected~\cite{feichtenhofer_masked_2022}. As part of the data augmentations for pre-training, we employ random resized crops~\cite{szegedy_going_2014} in spatial domain, random scale $\in [0.5, 1]$, and random horizontal flipping~\cite{feichtenhofer_masked_2022}.

\paragraph{Network architecture.} For our experiments we use vanilla ViT ~\cite{dosovitskiy_image_2021}. Specifically, we select ViT-Base to save GPU memory and avoid longer pre-training time. We use patch size of $2 \times 3 \times 16 \times 16$~\cite{arnab_vivit_2021, tong_videomae_2022, feichtenhofer_masked_2022}, resulting in $(16/2) \times (3/3) \times (224/16) \times (224/16) = 1568$ tokens for an input video of size $16 \times 3 \times 224 \times 224$. See supplementary material for more network architecture details.

\paragraph{Pre-training setting.} For all experiments, the default number of pre-training epochs is 800 unless otherwise noted. We use \texttt{adamw}~\cite{loshchilov_decoupled_2019} optimizer with a batch size of 32/GPU and 8 GPUs. See supplementary material for more details.

\paragraph{Evaluation on action classification.} \label{sec:evaluation} In order to evaluate the quality of the pre-trained model, we perform end-to-end fine-tuning (instead of linear probing~\cite{bao_beit_2022, he_masked_2021, tong_videomae_2022, feichtenhofer_masked_2022}). During  inference, we follow the common practice of multi-view testing~\cite{wang_non-local_2018,neimark_video_2021} where $K$ temporal clips (by default $K=2$~\cite{tong_videomae_2022} and 7~\cite{tong_videomae_2022, feichtenhofer_masked_2022} on SSv2 and Kinetics, respectively) to cover the video length.  For each clip, we take 3 spatial views~\cite{feichtenhofer_masked_2022, tong_videomae_2022} to cover the complete image. The final prediction is the average across all views.

%%%%%%%%%%%%%%%%%%%%%%%%%%%%%%%%%%%%%%%%%%%%%%%%%%%%%%%%%%%%%%%
%%%%%%%%%%%%%%%%%%%%%% Results and Discussion %%%%%%%%%%%%%%%%%
%%%%%%%%%%%%%%%%%%%%%%%%%%%%%%%%%%%%%%%%%%%%%%%%%%%%%%%%%%%%%%%
\section{Results and Discussion}
%%%%%%%%%%%%%%%%%%%%%%%%%%%%%%%%%%
%%%%%% Ablation Experiments %%%%%%
%%%%%%%%%%%%%%%%%%%%%%%%%%%%%%%%%%
\subsection{Ablation studies}
We perform pre-training using \adamae on ViT-Base backbone and then fine-tune the encoder with supervision for evaluation on the SSV2 dataset. We present our ablation study results in Table \ref{tab:ablations}. In the following sections, by accuracy we  refer to top-1 accuracy unless stated otherwise. 

% Ablation table
\input{latex/tables/ablations}
    
% Making technique
\paragraph{Adaptive \textit{vs.} random patch, frame, and tube masking.} We compare different masking techniques (as illustrated in Fig. \ref{fig:mask_type_comp}) in Table \ref{tab:mask_types}. Random \textit{tube masking}~\cite{tong_videomae_2022}, which randomly samples masked tokens in the 2D spatial domain and then extends those along the temporal axis, works reasonably well with a high masking ratio (69.3\% accuracy with 90\% masking). Random \textit{patch masking}~\cite{feichtenhofer_masked_2022}, which randomly masks tokens over spacetime, also works well with high masking ratios (68.3\% accuracy with 90\% masking \textit{vs.} 67.3\% accuracy with 75\% masking) - note that we observe 1\% drop in performance compared to random \textit{tube masking}~\cite{tong_videomae_2022}. However, random \textit{frame masking}, which masks out all tokens in randomly selected frames, performs poorly (61.5\% accuracy) compared to random patch and tube masking. Our \textbf{\textit{adaptive token sampling}}, which samples visible tokens based on a categorical distribution (instead of assuming a fixed distribution) estimated from a neural network by exploiting the spatiotemporal relationship of all tokens, yields the \textit{best} result (70.04\% accuracy) -- \textbf{~0.7\%} improvement over random  patch and tube masking. Note that this improvement is achieved with a higher masking ratio (95\%) which requires less memory (14.4 GB \textit{vs.} 14.7 GB) and results in shorter pre-training time.

% Pred mask visualization
\input{latex/figs/pred_mask}
% Masking ratio
\paragraph{Masking ratio.} Table \ref{tab:masking_ratio} shows the performance of our model while using different masking ratios. Our \textbf{\textit{adaptive token sampling}} achieves the \textbf{best} performance for masking ratio of \textbf{95\%} (70.04\% accuracy); 5\% higher masking ratio compared to random \textit{patch masking} in  SpatiotemporalMAE~\cite{feichtenhofer_masked_2022} (90\% masking) and random \textit{tube masking} in VideoMAE~\cite{tong_videomae_2022} (90\% masking). As shown in Fig. \ref{fig:mask85} and \ref{fig:mask95}, we sample a relatively higher number of tokens from regions with less redundancy and a small number of tokens from regions with high redundancy. Hence, fewer tokens are required than random sampling to achieve the same reconstruction error. 98\% and 90\% masking ratios also perform reasonably well (68.85\% and 69.55\% accuracies, respectively), whereas we observe a significant drop in performance for 85\% and 80\% masking ratios (68.06\% and 66.96\% accuracies, respectively). For lower masking ratios, due to large number of redundant visible patches being sampled, the network just copies features from these patches, resulting in poor generalizations and drop in performance. 
%This is mainly due to more redundancy in the input that results in a lazy network that mostly copies patches as opposed to learning useful features.
% Decoder depth
\paragraph{Decoder depth.} We compare our performance for different decoder depths (i.e., number of transformer blocks used) in Table \ref{tab:decoder_depth}. Increasing the decoder depth from 1 block to 4 blocks increases the accuracy (from 68.86\% to 70.04\%); however, this comes with an additional longer pre-training time. Increasing the decoder depth from 4 blocks to 8 blocks, however, results in a slight drop in accuracy (from 70.04\% to 69.97\%). Hence, we consider 4 transformer blocks as the optimal decoder depth for \adamae, on par with the observation made in VideoMAE~\cite{tong_videomae_2022} and SpatiotemporalMAE~\cite{feichtenhofer_masked_2022}.

% Reconstruction target
\paragraph{Reconstruction target.} Table~\ref{tab:loss_target} shows the impact of employing different loss functions and reconstruction targets. We use RGB pixel values as the reconstruction target; instead of utilizing complicated features such as HoG~\cite{wei_masked_2021}, which always comes with additional computational cost. Following an improved performance with per-patch normalized pixel values for images in ImageMAE~\cite{he_masked_2021} and also verified later for videos in VideoMAE~\cite{tong_videomae_2022} and SpatiotemporalMAE~\cite{feichtenhofer_masked_2022}, we consider two scenarios -- with and without local patch normalization. Regression over raw RGB pixels with the L1/MSE loss has shown poor performance compared to regression over per-patch normalized pixels. We observe that MSE loss has a slight advantage over L1 loss.

% Pre-training epochs
\paragraph{Pre-training epochs.} Next, we show in Table~\ref{tab:pretraining_epochs} the effect of the number of pre-training epochs on fine-tuning performance. Increasing the number of pre-training epochs always increases the performance: for 200 epochs - 66.42\%, for 500 epochs 69.20\%, for 600 epochs - 69.47\%, for 700 epochs 69.68\*, and for 800 epochs we achieve the best result of 70.04\%. We observe poor performance when pre-training for fewer than 500 epochs. Hence, considering the trade-off between pre-training time and performance, we recommend pre-training with \adamae for at least 500 epochs.

% Mask sampling network
\paragraph{Mask sampling network.} Finally, we investigate the effect of using different network architectures for our adaptive token sampling network in Table \ref{tab:adamae_network}. Considering the importance of a computationally lightweight design for an adaptive token sampling network, we first experimented with a simple multi-layer perceptron (MLP) network (see supplementary material).  The results of this MLP network were not encouraging, giving similar performance to random patch masking. We visualize the predicted probability map to confirm that it was uncorrelated to the video itself (see supplementary material). 
%as shown in Fig. \ref{fig:sampling_network}-(a). 
%Although this network has a lightweight design and brings only a little computational overhead, the experimental results were not encouraging. Specifically, it was unable to capture spatiotemporal information of the input tokens, resulting in the predicted probability density map having almost no relationship with the video. % as shown in Fig. \textbf{???? Did we omit this figure?} . 
%Therefore, the sampled visible tokens acted similar to a random sample resulting in no improvements over previous results. 
This is expected since a simple MLP network cannot model the relationship between the tokens as it operates only on the embedding dimension of each token. We next experimented with an  MHA network (see supplementary material). % (i.e., a transformer block) as shown in Fig. \ref{fig:sampling_network}-(b). 
This network can capture the spatiotemporal relationships between tokens through the attention mechanism and is able to predict meaningful probability density maps, as shown in Figs. \ref{fig:mask85}, and \ref{fig:mask95}. Increasing the number of MHA blocks results in slight improvements in performance, albeit with more memory and compute requirements. We selected a single MHA block as our adaptive token sampling  network for all other experiments to keep it as computationally lightweight as possible. Furthermore, we experimented with different embedding dimensions (i.e., $d$ and $d/2$) and observed improved performance when operating MHA at its original embedding dimension $d$.

%%%%%%%%%%%%%%%%%%%%%%%%%%%%%%%%%%%%%%%%%%%%%
%%%%%%%%%%      Main Analysis      %%%%%%%%%%
%%%%%%%%%%%%%%%%%%%%%%%%%%%%%%%%%%%%%%%%%%%%%
\subsection{Main results and analysis}
We compare the performance of \adamae with previous SOTA results for action classification on the  SSv2~\cite{goyal_something_2017} and Kinetics-400~\cite{kay_kinetics_2017} datasets. Our results are presented in Table \ref{tab:ssv2} (on SSv2) and Table \ref{tab:k400} (on Kinetics-400) for the ViT-Base backbone ($\sim$ 87M parameters). 
%Although we present the numbers for previous SOTA methods for very large backbones such as ViT-L and ViT-H in the comparisons, we \textcolor{gray}{discard} those in the final comparison.

MAEs (MaskFeat, VideoMAE, SpatioTemporalMAE, OmniMAE, and \adamae) generally outperform previous supervised representation learning approaches (such as TDN~\cite{wang_tdn_2021}, TimeSformer~\cite{bertasius_is_2021}, Motionformer~\cite{patrick_keeping_2021}, Video-Swin~\cite{liu_video_2021}),  which use ImageNet-1K (IN1K), ImageNet-21K (IN21K), Kinetics-400, and/or Kinetics-600 labels for supervision, by a significant margin on these datasets. This empirically demonstrates the powerful representation learning capability of masked reconstruction methods. Furthermore, they also outperform recent contrastive learning approaches such as BEVT~\cite{wang_bevt_2022} and hybrid approaches (i.e., masked modeling and contrastive learning) such as VIMPAC~\cite{tan_vimpac_2021}. 

We compare \adamae with recent MAE-based spatiotemporal representation learning approaches: VideoMAE~\cite{tong_videomae_2022}, SpatioTemporalMAE~\cite{feichtenhofer_masked_2022}, and OmniMAE~\cite{girdhar_omnimae_2022} with the same ViT-Base backbone. Among these,  VideoMAE performs best when pre-trained with 90\% random tube masking (69.3\% and 80.9\% in accuracies on SSv2 and Kinetics-400, respectively). SpatioTemporalMAE achieves its best performance with 90\% random patch masking (68.3\% and 81.3\%  accuracies on SSv2 and Kinetics-400, respectively). OminiMAE shows lower performance than VideoMAE and SpatiotemporalMAE on  SSv2 and Kinetics-400 datasets (69.3\% and 80.6\% accuracies on SSv2 and Kinetics-400, respectively) with 95\% random patch masking, even though it uses extra data (IN1K with 90\% random patch masking, and Kinetics-400 and SSv2 with 95\% random masking). In contrast, our AdaptiveMAE, which utilizes an adaptive masking scheme based on the spatiotemporal information of a given video, achieves the best performance on the SSv2 and Kinetics-400 datasets (70.0\% and 81.7\% accuracies, respectively) with higher masking ratio (95\%). This results in faster pre-training compared to VideoMAE and SpatitemporalMAE and requires less GPU memory.

\paragraph{Transferability:} We investigate the transferability of \adamae pre-trained models by evaluating their performance in scenarios where the pre-training and finetuning datasets are different. Our \adamae \textit{pre-trained on Kinetics-400} achieves SOTA results of \textbf{69.9\%} (top-1 accuracy) and 92.7\% (top-5 accuracy) on \textit{SSv2}. On the other hand, \adamae \textit{pre-trained on SSv2} achieves \textbf{81.2\%} (top-1 accuracy) and 94.5\% (top-5 accuracy) on \textit{Kinetics-400}, which is better than other frameworks.

%%%% Main results SSv2 %%%
\input{latex/tables/results_ssv2}

%%%% Main results K400 %%%%
\input{latex/tables/results_k400}

%%%%%%%%%%%%%%%%%%
%% Limitations %%%
%%%%%%%%%%%%%%%%%%
\section{Limitations and Future Work}
Although we have presented results for ViT-Base backbone, our approach for adaptive token sampling scales-up for larger backbones (e.g., \textit{ViT-Large} and \textit{ViT-Huge}) as well. We expect \adamae to achieve better results for these settings based on the experiments in previous studies~\cite{tong_videomae_2022, feichtenhofer_masked_2022}, and we plan to conduct similar studies in future. Since \adamae outputs a categorical distribution on the tokens, it could be used for applications such as saliency detection in videos. Moreover, \adamae needs only 5\% of the video tokens to represent the entire video reasonably well. Hence, it could be applied for efficient video compression and retrieval.

%%%%%%%%%%%%%%%%%%%%
%%%% Conclusion %%%%
%%%%%%%%%%%%%%%%%%%%
\section{Conclusion}
We propose an adaptive masking technique for MAEs, \adamae, that efficiently learns meaningful spatiotemporal representations from videos for downstream tasks. Unlike existing masking methods such as random patch, tube, and frame masking, \adamae samples visible tokens based on a categorical distribution. An auxiliary network is optimized to learn the distribution by maximizing the expected reconstruction error using policy gradients. \adamae outperforms previous state of the art methods with ViT-Base model on Something-Something v2 and Kinetics-400 datasets and learns better transferable features, while lowering GPU memory requirements and masking pre-training faster.

%%%% Appendix %%%%
\appendix
\input{latex/appendix}

%%%%%%%%%%%%%%%%%%
%%% REFERENCES %%%
%%%%%%%%%%%%%%%%%%
\clearpage
{\small
\bibliographystyle{ieee_fullname}
\bibliography{references}
}

\end{document}

%% file: latex/tables/ablations.tex
\begin{table*}[tbh]
    \vspace{-.2em}
    \centering
    % Masking types
    \subfloat[
    \textbf{Mask sampling techniques}. Our \textit{adaptive} masking works better compared to tube, frame, and random, and requires less memory.
    \label{tab:mask_types}
    ]{
    \begin{minipage}{0.34\linewidth}{\begin{center}
    \tablestyle{4pt}{1.05}
    \begin{tabular}{y{50}x{20}x{20}x{20}x{28}}
    case & ratio & top-1 & top-5 & memory\\
    \shline
    Tube~\cite{tong_videomae_2022} & 75 & 67.3 & 91.5 & 16.2 GB\\
    Tube~\cite{tong_videomae_2022} & 90 & 69.3 &  92.3 & 14.7 GB\\
    Random~\cite{feichtenhofer_masked_2022} & 75 & 67.9 & 90.6 & 16.2 GB\\
    Random~\cite{feichtenhofer_masked_2022} & 90 & 68.3 & 91.8  & 14.7 GB\\
    Frame~\cite{feichtenhofer_masked_2022, tong_videomae_2022} & 87.5 &  61.5 & 87.6 & 12.8 GB\\
    \baseline{Adaptive (ours)} & \baseline{95} & \baseline{\textbf{70.04}} & \baseline{\textbf{92.70}} & \baseline{14.4 GB}\\
    \end{tabular}
    \end{center}}\end{minipage}
    }
    \hspace{2em}
    % Masking percentage.
    \subfloat[
    \textbf{Different masking ratio $(\rho)$}. Our AdaMAE works well with extremely high masking ratio, hence requires less memory.
    \label{tab:masking_ratio}
    ]{
    \begin{minipage}{0.27\linewidth}{\begin{center}
    \tablestyle{4pt}{1.05}
    \begin{tabular}{x{24}x{22}x{22}x{32}}
    ratio &  top-1 & top-5 & memory\\
    \shline
    0.98 & 68.85& 91.94 & 14.1 GB\\
    \baseline{0.95} & \baseline{\textbf{70.04}} & \baseline{\textbf{92.70}} & \baseline{14.4 GB}\\
    0.90 & 69.55& 92.62 & 15.1 GB\\
    0.85 & 68.06& 91.38 & 15.9 GB\\
    0.80 & 66.96& 90.91 & 16.9 GB\\
    \multicolumn{3}{c}{~} \\
    \end{tabular}
    \end{center}}\end{minipage}
    }
    \hspace{2em}
    % Decoder depth
    \subfloat[
    \textbf{Different decoder depth $(D)$}. Our AdaMAE achives the best performance with 4 blocks of decoder.
    \label{tab:decoder_depth}
    ]{
    \centering
    \begin{minipage}{0.24\linewidth}{\begin{center}
    \tablestyle{4pt}{1.05}
    \begin{tabular}{x{18}x{24}x{24}x{28}}
    blocks & top-1 & top-5 & memory\\
    \shline
    1 &  68.86 & 92.07 & 9.2 GB\\ 
    2 &  69.13 & 92.17 & 11.0 GB\\
    \baseline{4} & \baseline{\textbf{70.04}} & \baseline{\textbf{92.70}} & \baseline{14.4 GB}\\
    8 & 69.97 & 92.67 & 21.3 GB\\
    \multicolumn{3}{c}{~}\\
    \multicolumn{3}{c}{~}\\
    \end{tabular}
    \end{center}}\end{minipage}
    }
\\
    \centering
    \vspace{.5em}
    %Loss functions
    \subfloat[
    \textbf{Loss function}. MSE loss with local patch normalization gives the best results.
    \label{tab:loss_target}
    ]{
    \begin{minipage}{0.29\linewidth}{\begin{center}
    \tablestyle{4pt}{1.05}
    \begin{tabular}{y{72}x{24}x{24}}
    case & top-1 & top-5 \\
    \shline
    L1 loss (w norm.)  &  69.12 & 91.42 \\
    L1 loss (w/o norm.) & 68.75 & 91.47 \\
    \baseline{MSE loss (w norm.)}  &  \baseline{\textbf{70.04}} & \baseline{\textbf{92.70}} \\
    MSE loss (w/o norm.) & 68.87 & 91.45 \\
    \end{tabular}
    \end{center}}\end{minipage}
    }
    \hspace{2em}
    %Pre-training epochs
    \subfloat[
    \textbf{Pre-training epochs}. Better performance with a greater number of pre-training epochs.
    \label{tab:pretraining_epochs}
    ]{
    \centering
    \begin{minipage}{0.22\linewidth}{\begin{center}
    \tablestyle{4pt}{1.05}
    \begin{tabular}{x{20}x{20}x{22}}
    epochs & top-1 & top-5 \\
    \shline
    200 & 66.42 & 90.59\\
    500 & 69.20 & 92.50\\
    600 & 69.47 & 92.52\\
    \baseline{800} & \baseline{\textbf{70.04}} & \baseline{\textbf{92.70}}\\
    \end{tabular}
    \end{center}}\end{minipage}
    }
    \hspace{2em}
    % Mask sampling network
    \subfloat[
    \textbf{Mask sampling network}. Our AdaMAE works well with a single MHA. 
    \label{tab:adamae_network}
    ]{
    \centering
    \begin{minipage}{0.36\linewidth}{\begin{center}
    \tablestyle{4pt}{1.05}
    \begin{tabular}{y{82}x{18}x{18}x{32}}
    case & top-1 & top-5 & memory\\
    \shline
    MLP   & 65.41 & 89.22 & 11.8 GB\\
    MHA $(D = 1, d = 384)$ & 69.56 & 90.89 & 14.0 GB\\
    \baseline{MHA $(D = 1, d = 768)$} & \baseline{\textbf{70.04}} & \baseline{\textbf{92.70}} & \baseline{14.4 GB}\\
    MHA $(D = 2, d = 384)$ & 70.03 & 92.74 & 17.1 GB\\
    \end{tabular}
    \end{center}}\end{minipage}
    }
    \vspace{-.7em}
    \caption{\textbf{Ablation experiments on SSv2 dataset}. We use ViT-Base~\cite{vaswani_attention_2017} as the backbone for all experiments. MHA $(D=2, d=384)$ denotes our adaptive token sampling network with a depth of two and embedding dimension of $384$.  All pre-trained models are evaluated based on the evaluation protocol described in Sec. \ref{sec:evaluation}. The default choice of our \adamae is highlighted in \colorbox{baselinecolor}{\bf gray} color. The GPU memory consumption is reported for a batch size of 16 on a single GPU.
    }
    \label{tab:ablations} 
    \vspace{-.99em}
    \end{table*}

%% file: latex/figs/pred_mask.tex
%##########################################################
    \begin{figure}[!htb]
        % \centering
        % \begin{subfigure}[b]{0.495\textwidth}
        %     \centering
        %     \includegraphics[width=\linewidth]{latex/imgs/SSL_videos-80.jpeg}
        %     \caption[]%
        %     {{\small Masking ratio $= 80\%$.}}    
        %     \label{fig:mask80}
        % \end{subfigure}
        % \hfill
        \begin{subfigure}[b]{0.495\textwidth}  
            \centering 
            \includegraphics[width=\linewidth]{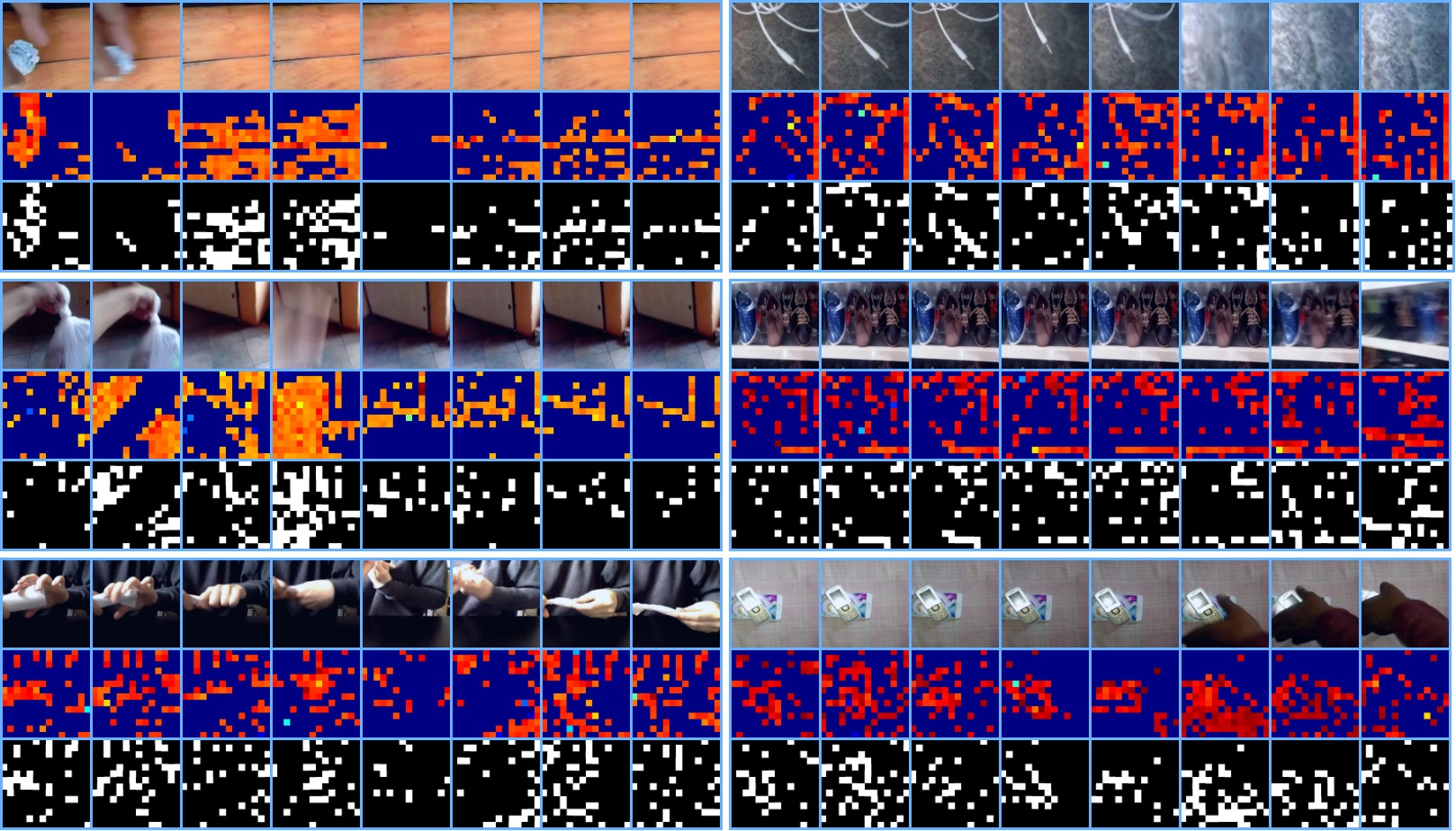}
            \caption[]%
            {{\small Masking ratio $= 85\%$.}} 
            \label{fig:mask85}
        \end{subfigure}
        \vskip\baselineskip
        % \begin{subfigure}[b]{0.495\textwidth}   
        %     \centering 
        %     \includegraphics[width=\linewidth]{latex/imgs/SSL_videos-90.jpeg}
        %     \caption[]%
        %     {{\small Masking ratio $= 90\%$.}} 
        %     \label{fig:mask90}
        % \end{subfigure}
        % \hfill
        \begin{subfigure}[b]{0.495\textwidth}   
            \centering 
            \includegraphics[width=\linewidth]{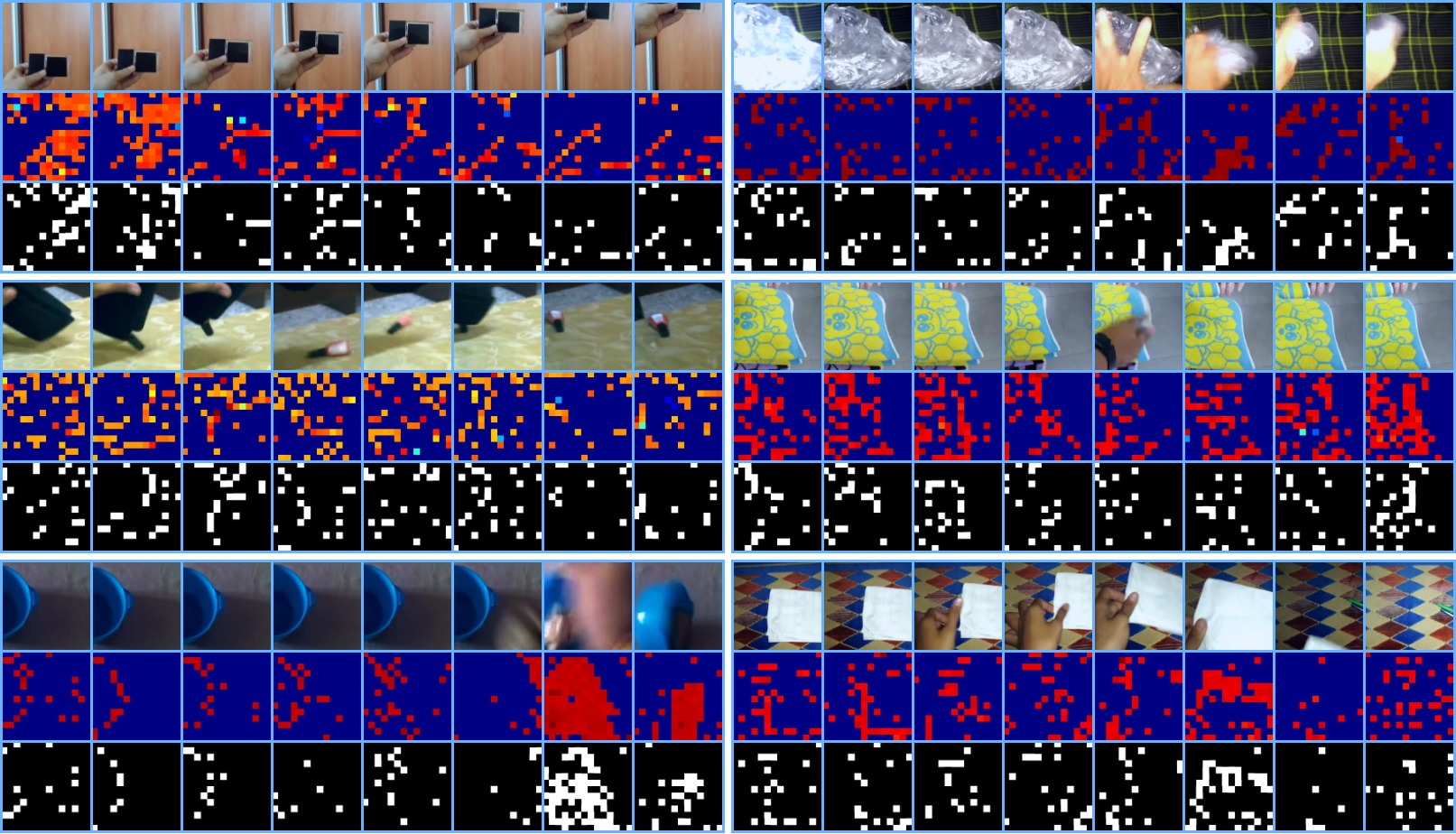}
            \caption[]%
            {{\small Masking ratio $= 90\%$.}} 
            \label{fig:mask95}
        \end{subfigure}
        \caption[ The average and standard deviation of critical parameters ]
        {\textbf{Mask visualizations of our \adamae for different masking ratios on SSV2 dataset~\cite{goyal_something_2017}.} 
        Given a video (\textit{first row}), our \adamae first predicts the categorical distribution (\textit{second row}), and then samples the mask (\textit{third row}) from that distribution. Refer to the supplementary material for more visualizations. %Colors closer \textcolor{red}{\bf red} and \textcolor{blue}{\bf blue} denotes the patches with \textit{high} and \textit{low} probability, respectively. In mask visualizations, \textbf{black} and \textbf{white} corresponds to the masked and visible path locations, respectively.
        }
        \vspace{-6mm}
        \label{fig:pred_masks}
    \end{figure}

%% file: latex/tables/results_ssv2.tex
\begin{table*}[t!]
    \centering
    \tablestyle{2.0pt}{1.04}
    \begin{tabular}{l|c|c|c|c|c|c|c|c|c}
    \textbf{Method} & \textbf{Backbone}  &\textbf{Extra data}  &\textbf{Extra labels} & \textbf{Pre. Epochs} & \textbf{Frames} & \textbf{GFLOPs}  & \textbf{Param} & \textbf{Top-1}  & \textbf{Top-5} \\
    
    \shline\hline
    TSM$_{two\ stream}$~\cite{lin_tsm_2019}  & ResNet50\x2  & \multirow{4}{*}{IN1K}  & \cmark & N/A & 16+16 & 130\x2\x3 & 49 & 66.0 & 90.5 \\ 
    TEINet$_{En}$~\cite{liu_teinet_2019}  & ResNet50\x2  &  & \cmark & N/A & 8+16  & 99\x10\x3 & 50 & 66.6 & N/A \\ 
    TANet$_{En}$~\cite{liu_tam_2021} &  ResNet50\x2 &  & \cmark & N/A & 8+16 & 99\x2\x3 & 51 & 66.0 & 90.1 \\
    TDN$_{En}$~\cite{wang_tdn_2021} & ResNet101\x2 &   & \cmark & N/A & 8+16 & 198\x1\x3 & 88 & 69.6 & 92.2   \\

    \hline
    SlowFast~\cite{feichtenhofer_slowfast_2019} &  ResNet101 &  \multirow{2}{*}{Kinetics-400} & \cmark & N/A & 8+32 & 106\x1\x3 & 53 & 63.1 & 87.6 \\
    MViTv1~\cite{fan_multiscale_2021} & MViTv1-B &  & \cmark & N/A & 64 & 455\x1\x3 & 37 & 67.7 & 90.9  \\
    
    \hline
    MViTv1~\cite{fan_multiscale_2021} & MViTv1-B & \multirow{2}{*}{Kinetics-600} & \cmark & N/A & 32 & 170\x1\x3 & 37  & 67.8 & 91.3 \\
    MViTv1~\cite{fan_multiscale_2021} & MViTv1-B-24 &  & \cmark & N/A & 32 & 236\x1\x3 & 53  & 68.7 & 91.5 \\
    \hline
    
    TimeSformer~\cite{bertasius_is_2021} & ViT-B & \multirow{2}{*}{IN21K} & \cmark & N/A & 8 & 196\x1\x3 & 121 & 59.5 &  N/A \\
    \textcolor{gray}{TimeSformer~\cite{bertasius_is_2021}} & \textcolor{gray}{ViT-L} &  & \cmark & N/A & 64  & 5549\x1\x3 & 430 & 62.4 & N/A \\
    \hline
    
    ViViT FE~\cite{arnab_vivit_2021} & ViT-L & \multirow{4}{*}{\footnotesize IN21K+Kinetics-400} & \cmark & N/A & 32 & 995\x 4\x3 & N/A & 65.9 & 89.9 \\
    Motionformer~\cite{patrick_keeping_2021} & ViT-B &   & \cmark & N/A & 16 & 370\x1\x3 & 109  & 66.5 & 90.1 \\
    Motionformer~\cite{patrick_keeping_2021} & ViT-L &   & \cmark & N/A & 32 & 1185\x1\x3 & 382  & 68.1 & 91.2 \\
    Video Swin~\cite{liu_video_2021}  & Swin-B &  & \cmark & N/A & 32 & 321\x1\x3 & 88 & 69.6 & 92.7  \\
    \hline
    
    VIMPAC~\cite{tan_vimpac_2021} & ViT-L & \scriptsize{HowTo100M+DALLE} & \xmark & N/A & 10 & N/A\x10\x3 & 307 & 68.1 & N/A \\
    BEVT-V~\cite{wang_bevt_2022}  & Swin-B & \scriptsize{Kinetics-400+DALLE} & \xmark & N/A & 32 & 321\x1\x3 & 88 & 67.1 & N/A  \\
    \textcolor{gray}{BEVT~\cite{wang_bevt_2022}}  & \textcolor{gray}{Swin-B} & \textcolor{gray}{\scriptsize{IN-1K+Kinetics-400+DALLE}} &\textcolor{gray}{\xmark}  & N/A & \textcolor{gray}{32} & \textcolor{gray}{321\x1\x3} & \textcolor{gray}{88} & \textcolor{gray}{70.6} & \textcolor{gray}{N/A}  \\
    
    \hline
    \textcolor{gray}{MaskFeat{\scriptsize\textuparrow312}~\cite{wei_masked_2021}} & \textcolor{gray}{MViT-L} & \textcolor{gray}{Kinetics-400}  &  \textcolor{gray}{\cmark} & N/A & \textcolor{gray}{40}  & \textcolor{gray}{2828\x1\x3} & \textcolor{gray}{218} & \textcolor{gray}{74.4} & \textcolor{gray}{94.6} \\
    \textcolor{gray}{MaskFeat{\scriptsize\textuparrow312}~\cite{wei_masked_2021}} & \textcolor{gray}{MViT-L} & \textcolor{gray}{Kinetics-600}  & \textcolor{gray}{\cmark}  & N/A & \textcolor{gray}{40}  & \textcolor{gray}{2828\x1\x3} & \textcolor{gray}{218} & \textcolor{gray}{75.0} & \textcolor{gray}{95.0} \\
    
    \hline
    MTV (B/2+S/4+Ti/8)~\cite{yan_multiview_2022}  & ViT-B & Kinetics-400+I21K & \cmark & N/A  & 32 & $384 \times 4 \times 3$ & 310 & 67.6 & 90.1\\ 
    
    \hline
    SpatioTemporalMAE~\cite{feichtenhofer_masked_2022} & ViT-B  & \emph{no external data} & \xmark & 800 & 16 & 180\x2\x3 & 87 & 68.3 & 91.8 \\
    %SpatioTemporalMAE~\cite{feichtenhofer_masked_2022} & ViT-L &  & \xmark & 16 & 597\x2\x3 & 305 &  &  \\
    %SpatioTemporalMAE~\cite{feichtenhofer_masked_2022} & ViT-L &  & \xmark & 32 & 1436\x1\x3 & 305 & \textbf{} & \textbf{} \\
    
    \hline
    VideoMAE~\cite{tong_videomae_2022} & ViT-B  & \emph{no external data} & \xmark & 800 & 16 & 180\x2\x3 & 87 & 69.3 & 92.3 \\
    %VideoMAE~\cite{tong_videomae_2022} & ViT-L &  & \xmark & 16 & 597\x2\x3 & 305 & 74.2 & 94.7 \\
    %VideoMAE~\cite{tong_videomae_2022} & ViT-L &  & \xmark & 32 & 1436\x1\x3 & 305 & 75.3 & 95.2 \\
    
    \hline
    OmniMAE~\cite{girdhar_omnimae_2022} & ViT-B & IN1K  & \xmark & 800 & 16 & 180\x5\x3 & 87 & 69.3 & N/A \\
    
    \hline
    \baseline{\bf \adamae$_{\rho=95\%}$ (ours)} & \baseline{ViT-B}    & \baseline{\emph{no external data}}  & \baseline{\xmark}  & \baseline{800}  & \baseline{16} & \baseline{180\x2\x3}    & \baseline{87}     & \baseline{\bf 70.0}     & \baseline{\bf 92.7} \\
    
    \shline\hline
    \end{tabular}
    \vspace{-2.5mm}
    \caption{\textbf{Comparison of our \adamae with SOTA methods on SSv2}~\cite{goyal_something_2017}. We report the results for ViT-Base~\cite{vaswani_attention_2017} architecture. Our model is pre-trained for the default setting in Table \ref{tab:ablations}. The ~\cmark~ in extra labels tab denotes supervised data used for pre-training while ~\xmark~ denotes only unlabeled data is used for the pre-training. The N/A denotes these numbers are not available/reported in the paper.}
    \vspace{-1mm}
    \label{tab:ssv2}
\end{table*}

%% file: latex/tables/results_k400.tex
%##########################################################
\begin{table*}[t!]
\centering
\tablestyle{2.0pt}{1.04}
\begin{tabular}{l|c|c|c|c|c|c|c|c|c}
    \textbf{Method} & \textbf{Backbone} & \textbf{Extra data} & \textbf{Extra labels} & \textbf{Pre. Epochs} &\textbf{Frames} & \textbf{GFLOPs}  & \textbf{Param} & \textbf{Top-1}  & \textbf{Top-5} \\
    
    % ResNet Backbones
    \shline\hline
    NL I3D~\cite{wang_non-local_2018} & ResNet101 &   \multirow{4}{*}{IN1K}    & \cmark & N/A &128 & 359\x10\x3  & 62  & 77.3 & 93.3    \\
    TANet~\cite{liu_tam_2021} &  ResNet152 & & \cmark & N/A & 16 & 242\x4\x3 & 59 & 79.3 & 94.1 \\
    TDN$_{En}$~\cite{wang_tdn_2021} & ResNet101$_{\times 2}$ &   & \cmark & N/A & 8+16 & 198\x10\x3 & 88 & 79.4 & 94.4   \\
    Video Swin~\cite{liu_video_2021}& Swin-B & & \cmark & N/A &32 & 282\x4\x3 & 88 & 80.6 & 94.6  \\
    
    % Slowfast
    \hline
    TimeSformer~\cite{bertasius_is_2021} & ViT-B & \multirow{6}{*}{IN21K} & \cmark & N/A & 8 & 196\x1\x3 & 121 & 78.3 &  93.7 \\
    TimeSformer~\cite{bertasius_is_2021} & ViT-L &  & \cmark & N/A & 96 & 8353\x1\x3 & 430 & 80.7 & 94.7 \\
    ViViT FE~\cite{arnab_vivit_2021} & ViT-L &  & \cmark & N/A & 128 & 3980\x 1\x3 & N/A & 81.7 & 93.8 \\
    Motionformer~\cite{patrick_keeping_2021} & ViT-B &    & \cmark & N/A & 16 & 370\x10\x3 & 109  & 79.7 & 94.2 \\
    Motionformer~\cite{patrick_keeping_2021} & ViT-L &    & \cmark & N/A & 32 & 1185\x10\x3 & 382  & 80.2 & 94.8 \\
    
    \textcolor{gray}{Video Swin~\cite{liu_video_2021}}  & \textcolor{gray}{Swin-L} & \textcolor{gray}{}  & \textcolor{gray}{\cmark} & \textcolor{gray}{N/A} & \textcolor{gray}{32} & \textcolor{gray}{604\x4\x3} & \textcolor{gray}{197} & \textcolor{gray}{83.1} & \textcolor{gray}{95.9}  \\
    
    %Video-VIT
    \hline
    \textcolor{gray}{ViViT FE~\cite{arnab_vivit_2021}} & \textcolor{gray}{ViT-L} & \textcolor{gray}{JFT-300M}  & \textcolor{gray}{\cmark} & \textcolor{gray}{N/A} & \textcolor{gray}{128} & \textcolor{gray}{3980\x 1\x3}  & \textcolor{gray}{N/A} & \textcolor{gray}{83.5} & \textcolor{gray}{94.3} \\
    
    ViViT~\cite{arnab_vivit_2021} & ViT-H & JFT-300M & \cmark & N/A & 32 & 3981\x 4\x3 & N/A & 84.9 & 95.8 \\
    
    VIMPAC~\cite{tan_vimpac_2021} & ViT-L & \scriptsize{HowTo100M+DALLE} & \xmark & N/A & 10 & N/A\x10\x3 & 307 & 77.4 & N/A \\
    
    BEVT~\cite{wang_bevt_2022}  & Swin-B & IN1K+DALLE  & \xmark & N/A & 32 & 282\x4\x3 & 88 & 80.6 & N/A  \\
    
    \hline
    ip-CSN~\cite{} & ResNet152 & \multirow{4}{*}{\emph{no external data}} & \xmark & N/A & 32 & 109$\times$10$\times$3 & 33 & 77.8 & 92.8 \\
    SlowFast~\cite{feichtenhofer_slowfast_2019} &  R101+NL &   & \xmark & N/A & 16+64 & 234\x10\x3 & 60 & 79.8 & 93.9 \\
    MViTv1~\cite{fan_multiscale_2021} & MViTv1-B &  & \xmark & N/A & 32 & 170\x5\x1 & 37  & 80.2 & 94.4 \\
    
    % MaskFeat
    \hline
    \textcolor{gray}{MaskFeat~\cite{wei_masked_2021}} & \textcolor{gray}{MViT-L} & \textcolor{gray}{} & \textcolor{gray}{\xmark} & \textcolor{gray}{N/A} & \textcolor{gray}{16}  & \textcolor{gray}{377\x10\x1} & \textcolor{gray}{218} & \textcolor{gray}{84.3} & \textcolor{gray}{96.3} \\
    
    \textcolor{gray}{MaskFeat{\scriptsize\textuparrow352}~\cite{wei_masked_2021}} & \textcolor{gray}{MViT-L} &  \textcolor{gray}{Kinetics-600}  & \textcolor{gray}{\xmark} & \textcolor{gray}{N/A} & \textcolor{gray}{40} & \textcolor{gray}{3790\x4\x3} & \textcolor{gray}{218} & \textcolor{gray}{87.0} & \textcolor{gray}{97.4} \\
    
    % VideoMAE
    \hline
    VideoMAE~\cite{tong_videomae_2022} & ViT-B & \multirow{3}{*}{\emph{no external data}} & \xmark & 1600 &16 & 180\x5\x3 & 87 & 80.9 & 94.7 \\
    
    \textcolor{gray}{VideoMAE~\cite{tong_videomae_2022}} & \textcolor{gray}{ViT-L} & & \textcolor{gray}{\xmark} & \textcolor{gray}{1600} & \textcolor{gray}{16} & \textcolor{gray}{597\x5\x3} & \textcolor{gray}{305} & \textcolor{gray}{84.7} & \textcolor{gray}{96.5} \\
    
    \textcolor{gray}{VideoMAE{\scriptsize\textuparrow320}~\cite{tong_videomae_2022}} & \textcolor{gray}{ViT-L} & & \textcolor{gray}{\xmark} & \textcolor{gray}{1600} & \textcolor{gray}{16} & \textcolor{gray}{3958\x5\x3} & \textcolor{gray}{305} & \textcolor{gray}{85.8} & \textcolor{gray}{97.1} \\
    
    %Spatiotemporal MAE
    \hline
    SpatioTemporalMAE~\cite{feichtenhofer_masked_2022} & ViT-B & \multirow{3}{*}{\emph{no external data}} & \xmark & 1600 &16 & 180\x3\x7 & 87 & 81.3  & 94.9\\
    
    \textcolor{gray}{SpatioTemporalMAE~\cite{feichtenhofer_masked_2022}} & \textcolor{gray}{ViT-L} & & \textcolor{gray}{\xmark} & \textcolor{gray}{1600} & \textcolor{gray}{16} & \textcolor{gray}{598\x3\x7} & \textcolor{gray}{305} & \textcolor{gray}{84.8}  & \textcolor{gray}{96.2} \\
    
    \textcolor{gray}{SpatioTemporalMAE~\cite{feichtenhofer_masked_2022}} & \textcolor{gray}{ViT-H} & \textcolor{gray}{} & \textcolor{gray}{\xmark} & \textcolor{gray}{1600} & \textcolor{gray}{16} & \textcolor{gray}{1193\x3\x7} & \textcolor{gray}{632} & \textcolor{gray}{85.1} & \textcolor{gray}{96.6}\\
    
    %Omni-MAE
    \hline
    OmniMAE~\cite{feichtenhofer_masked_2022} & ViT-B & \multirow{3}{*}{IN1K + SSv2} & \xmark & 1600 &16 & 180\x3\x7 & 87 & 80.6  & N/A\\
    
    \textcolor{gray}{OmniMAE{\scriptsize\textdownarrow512}~\cite{feichtenhofer_masked_2022}} & \textcolor{gray}{ViT-L} & & \textcolor{gray}{\xmark} & \textcolor{gray}{1600} & \textcolor{gray}{16} & \textcolor{gray}{598\x3\x7} & \textcolor{gray}{304} & \textcolor{gray}{84.0}  & \textcolor{gray}{N/A} \\
    
    \textcolor{gray}{OmniMAE~\cite{feichtenhofer_masked_2022}} & \textcolor{gray}{ViT-H} & \textcolor{gray}{} & \textcolor{gray}{\xmark} & \textcolor{gray}{1600} & \textcolor{gray}{16} & \textcolor{gray}{1193\x3\x7} & \textcolor{gray}{632} & \textcolor{gray}{84.8} & \textcolor{gray}{N/A}\\

    % AdaMAE (ours)
    \hline
    \baseline{\bf \adamae$_{\rho=95\%}$ (ours)} & 
    \baseline{ViT-B} & \baseline{\multirow{1}{*}{\emph{no external data}}} & 
    \baseline{\xmark} & 
    \baseline{800} & 
    \baseline{16} & 
    \baseline{180\x5\x3} & \baseline{87} & \baseline{\bf 81.7} & 
    \baseline{\bf 95.2} \\
    \shline\hline
\end{tabular}
\vspace{-0.5mm}
\caption{\textbf{Comparison of our \adamae with SOTA methods on Kinetics-400}~\cite{kay_kinetics_2017}. We report the results for ViT-Base~\cite{vaswani_attention_2017} architecture. Our model is pre-trained for the default setting in Table \ref{tab:ablations}. The ~\cmark~ in extra labels tab denotes supervised data used for pre-training while ~\xmark~ denotes only unlabeled data is used for the pre-training. The N/A denotes these numbers are not available/reported in the paper.}
\vspace{-12pt}
\label{tab:k400}
\end{table*}

%% file: latex/appendix.tex
\clearpage
\section{Supplementary material}
\subsection{Pseudocode for AdaMAE}
The pseudocode for our \texttt{AdaptiveTokenSampler} network is given in Algorithm \ref{alg:AdaptiveTokenSampling} and AdaMAE for \ref{alg:MAEPretraining}.
\begin{algorithm}[H]
        \begin{algorithmic}
            \State \textbf{Inputs:} Tokenized video: $\boldsymbol{X} \in \mathbb{R}^{N \times d}$, Masking ratio: $\rho \in (0, 1)$
            
            \State \textbf{Outputs:} Categorical distribution: $\boldsymbol{p}$, Sampled mask indices: $\boldsymbol{I}_m$, Sampled mask: $\boldsymbol{M}$
            
            \small
            \State $P = \texttt{TokenProbs}(\boldsymbol{X})$ \Comment{ token probabilities} 
            
            \State $\boldsymbol{p} \sim \text{CAT}(
            \boldsymbol{P})$ \Comment{ categorical distribution}
            
            \State $N_{\text{v}} = \texttt{int}(N \times (1-\rho))$ \Comment{ \# of visible tokens}
            
            \State $\boldsymbol{I}_v = \boldsymbol{p}.\texttt{sample}(N_{\text{vis}})$ \Comment{ visible token indices}
            
            \State $\boldsymbol{I}_m = \boldsymbol{U} - \boldsymbol{I}_v$ \Comment{ mask token indices}
            
            \State $\boldsymbol{M} = \texttt{GetBooleanMask}(\boldsymbol{I}_v, \boldsymbol{I}_m)$ \Comment{ binary mask}
        \end{algorithmic}
        \caption{Pseudo-code for \texttt{AdaptiveTokenSampler}.}
        \label{alg:AdaptiveTokenSampling}
\end{algorithm}

The pseudocode for our complete \texttt{AdaMAE} is given in \ref{alg:MAEPretraining} (for simplicity we assume a batch size of one).
\begin{algorithm}[H]
    \begin{algorithmic}
        \State \textbf{Input:} The video dataset $\mathcal{D}$ =  $\{\boldsymbol{V}_i: \text{ for } i=1,2,3, \cdots, |\mathcal{D}|\}$, Masking ratio: $\rho \in (0,1)$
        \For{$\boldsymbol{V}_i \in \mathcal{D}$}
            \small
            \State $\boldsymbol{X}_i =  \texttt{Tokenizer}(\boldsymbol{V}_i)$ \Comment{Tokens}
                
            \State $\boldsymbol{X}_i = \boldsymbol{X}_i + \texttt{PosEmbd}$ \Comment{Positional embedding}
                
            \State $\boldsymbol{p}, \boldsymbol{I}_v, \boldsymbol{M}$ $=\texttt{AdaptiveTokenSampler}(\boldsymbol{X}_i, \rho)$
                
            \State $\boldsymbol{X}_{v} = \boldsymbol{X}_{i}[\sim \boldsymbol{M}]$ \Comment{ visible tokens}
                
            \State $\boldsymbol{F}_{v} = \texttt{ViTEncoder}(X_{v})$ \Comment{ visible feats}
                
            \State $\boldsymbol{F}_{v} = \boldsymbol{F}_{v} + \texttt{PosEmbd}[\sim \boldsymbol{M}]$ 
                
            \State $\boldsymbol{F}_{m} = f_{m} + \texttt{PosEmbd}[\boldsymbol{M}]$
                \Comment{ mask feats}
                
            \State $\boldsymbol{F} = \boldsymbol{F}_{v} \oplus \boldsymbol{F}_{m} $ \Comment{ visible + mask feats}
                
            \State $\widehat{\boldsymbol{X}} = \texttt{ViTDecoder}(\boldsymbol{F})$ \Comment{Decoder}
            
            \State $\widehat{\boldsymbol{X}}_{m} = \widehat{\boldsymbol{X}}[\boldsymbol{M}]$ \Comment{ masked prediction}
            
            \State $\boldsymbol{X}_{m} = \boldsymbol{X}[\boldsymbol{M}]$ \Comment{ masked GT}
                
            \State $\mathcal{L}_{R} = \lVert \widehat{\boldsymbol{X}}_{m} - \boldsymbol{X}_{m} \rVert_2$ \Comment{ reconstruction loss}
            
            \State $\mathcal{L}_{S} = - \left( \boldsymbol{p}.\texttt{LogProb}( \boldsymbol{I}_m) \right) \cdot \mathcal{L}_{R}.detach()$ \Comment{ sampling loss}
                
            \State $\mathcal{L} = \mathcal{L}_{R} + 1e-4 \cdot \mathcal{L}_{S}$ \Comment{ final loss}
                
            \State $\mathcal{L}.backward()$ \Comment{ back-propagation}
        \EndFor
    \end{algorithmic}
    \caption{Pseudo-code for our \adamae.}
    \label{alg:MAEPretraining}
\end{algorithm}

\subsection{Network architectures for adaptive sampling network}
\begin{figure}[!htb]
        \centering
        \includegraphics[width=0.8\linewidth]{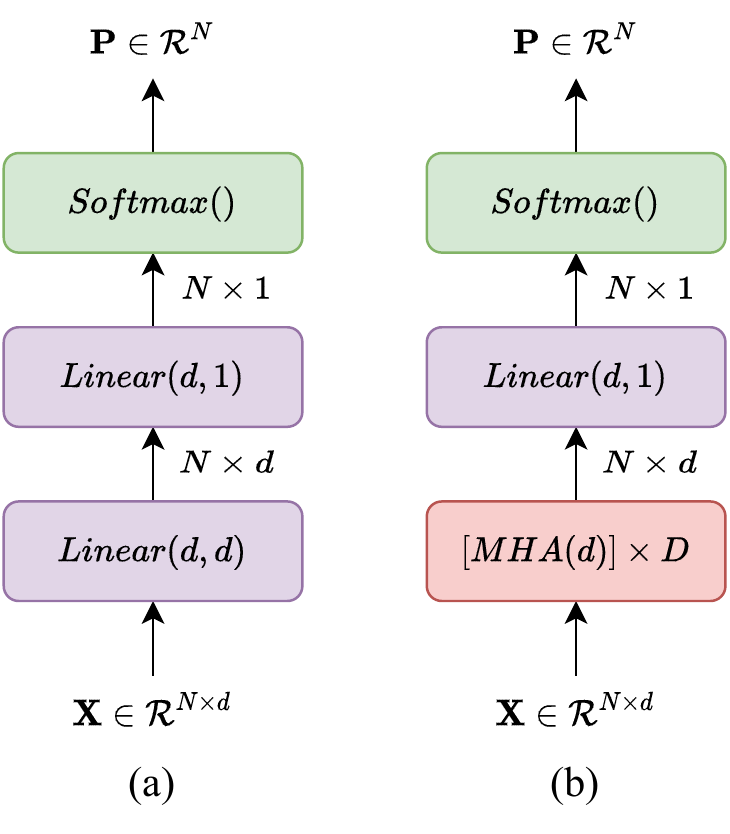}
        \caption{\textbf{The network architectures considered for adaptive token sampling network:} \textbf{(a)} MLP-based sampling network \textbf{(b)} MHA-based sampling network.}
        \label{fig:sampling_network}
    \end{figure}

    We consider two network architectures for our adaptive sampling network.
    \begin{enumerate}
        \item \textbf{MLP:} Keeping the importance of a computationally light-weight design for an adaptive sampling network, we first experimented with a simple MLP network for the sampling network as shown in Fig. \ref{fig:sampling_network} (a). In this architecture, we process all the tokens $(\mathbf{X} \in \mathbb{R}^{N \times d})$ through a $\texttt{Linear}$ layer with $\texttt{in\_features} = \texttt{out\_features} = d$ (denoted by $\texttt{Linear}(d, d)$) followed by a $\texttt{Linear} (d,1)$ to bring down the embedding dimension from $d$ to $1$. We then apply a $\texttt{Softmax}()$ layer to obtain the probability values $\mathbf{P} \in \mathbb{R}^N$.
        
        Although this network is computationally very light-weight, the experimental results were not encouraging. Specially, it was not able to capture the spatiotemporal information of input tokens; resulted in predicted probability density map to have no/very-less relationship with the high/low activity regions. This resulted in poor performance as demonstrated in Table \ref{tab:adamae_network}. This is understandable because it operates only on the embedding dimension and not be able to model the interaction between the patches that is necessary for predicting proper probability map.
        
        \item \textbf{MHA:} Motivated from the drawback that we observed with MLP-based sampling network, we decided to utilize multi-head self-attention network for this purpose, because it can properly model the interconnectivty between all the patches through the attention mechanism. Since we want to keep it computationally light-weight as much as possible, we experimented with the effect of different number of blocks and reduced emebdding demention as shown in Table \ref{tab:adamae_network}. By utilizing MHA-based sampling network, we were able to generate visually appeaing probability density maps (as shown in Figure \ref{fig:ss_loss_1}-\ref{fig:ss_loss_5}) that are in line with our intuition for adaptive sampling.
        
    \end{enumerate}

\subsection{Additional examples to understand the formulation of sampling loss $L_S$}
\label{sec:apex_ls_ad}
\begin{figure}[bh!]
    \centering
    \includegraphics[width=\linewidth]{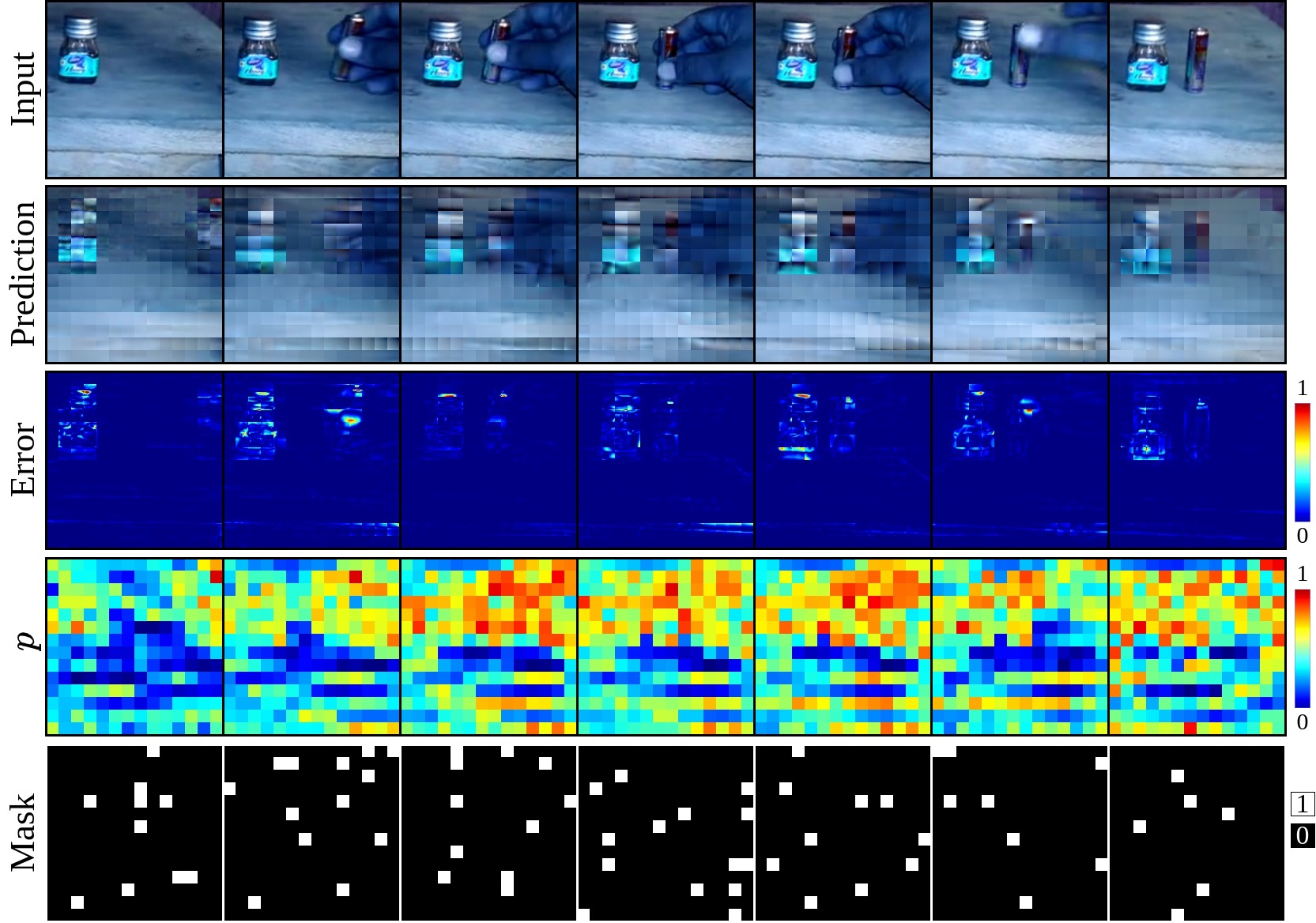}
    \caption{An example visualization of our adaptive sampling on SSv2 dataset.}
    \label{fig:ss_loss_1}
\end{figure}

\begin{figure}[tbh!]
    \centering
    \includegraphics[width=\linewidth]{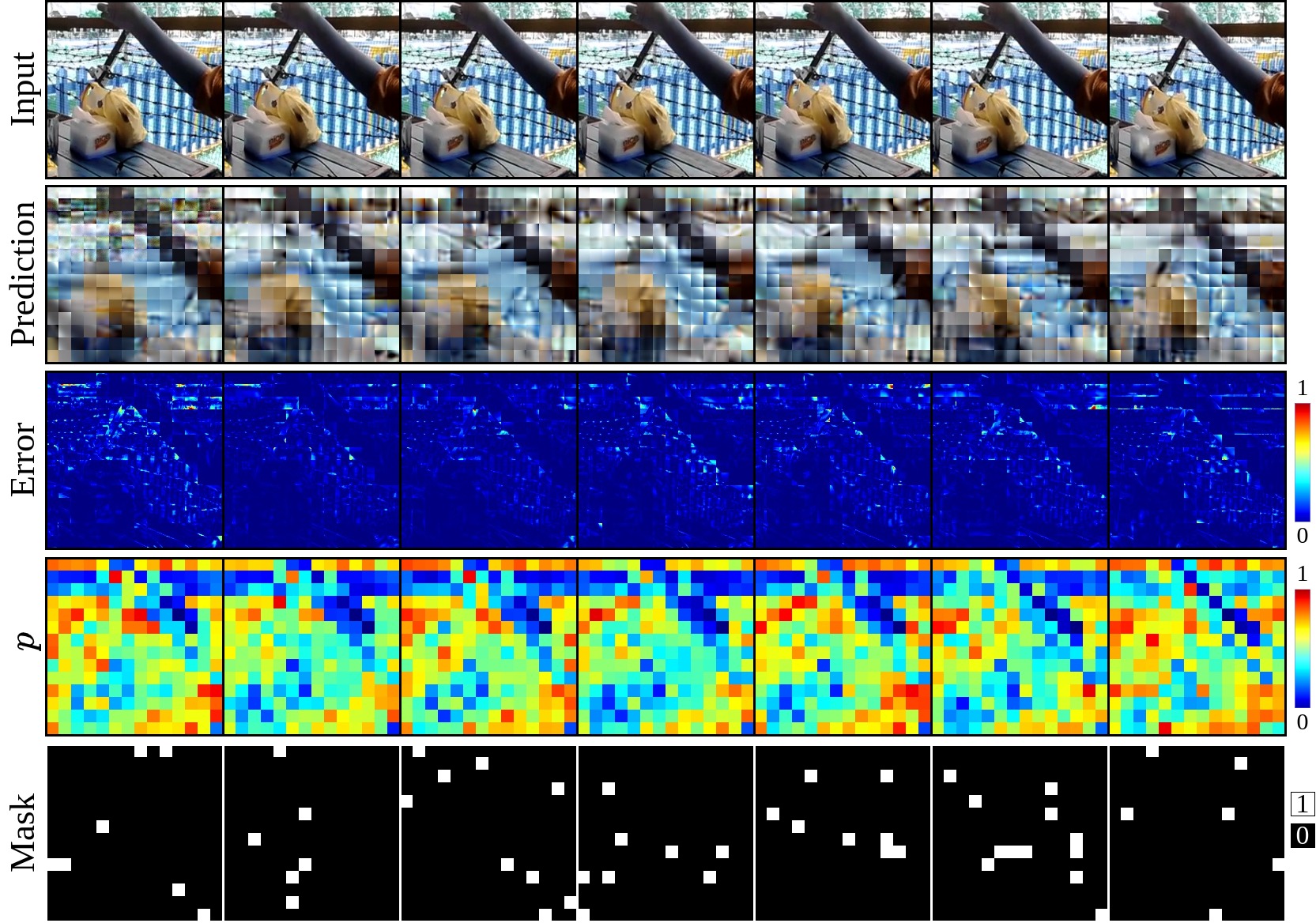}
    \caption{An example visualization of our adaptive sampling on SSv2 dataset.}
    \label{fig:ss_loss_2}
\end{figure}

\begin{figure}[tbh!]
    \centering
    \includegraphics[width=\linewidth]{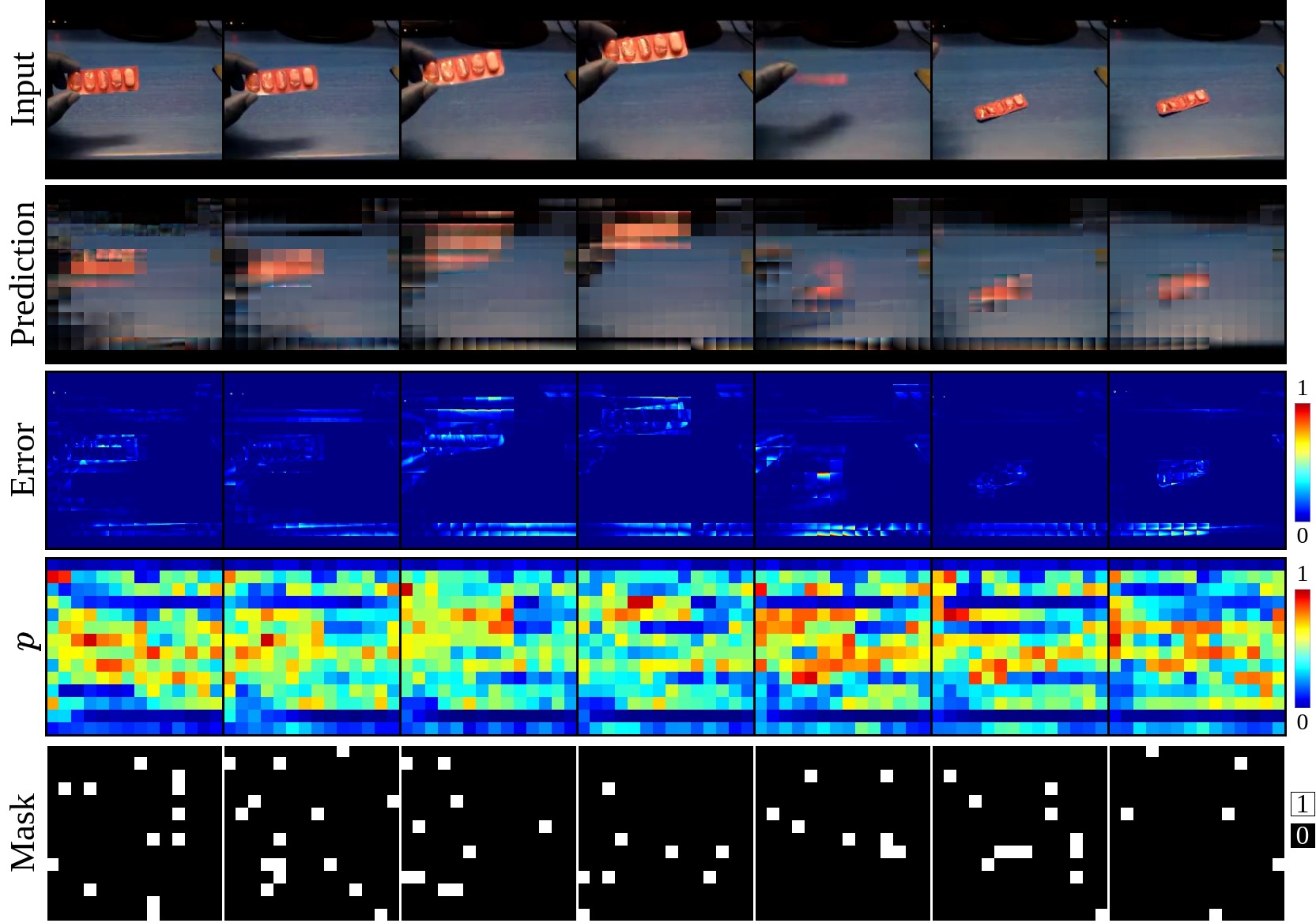}
    \caption{An example visualization of our adaptive sampling on SSv2 dataset.}
    \label{fig:ss_loss_3}
\end{figure}

\begin{figure}[tbh!]
    \centering
    \includegraphics[width=\linewidth]{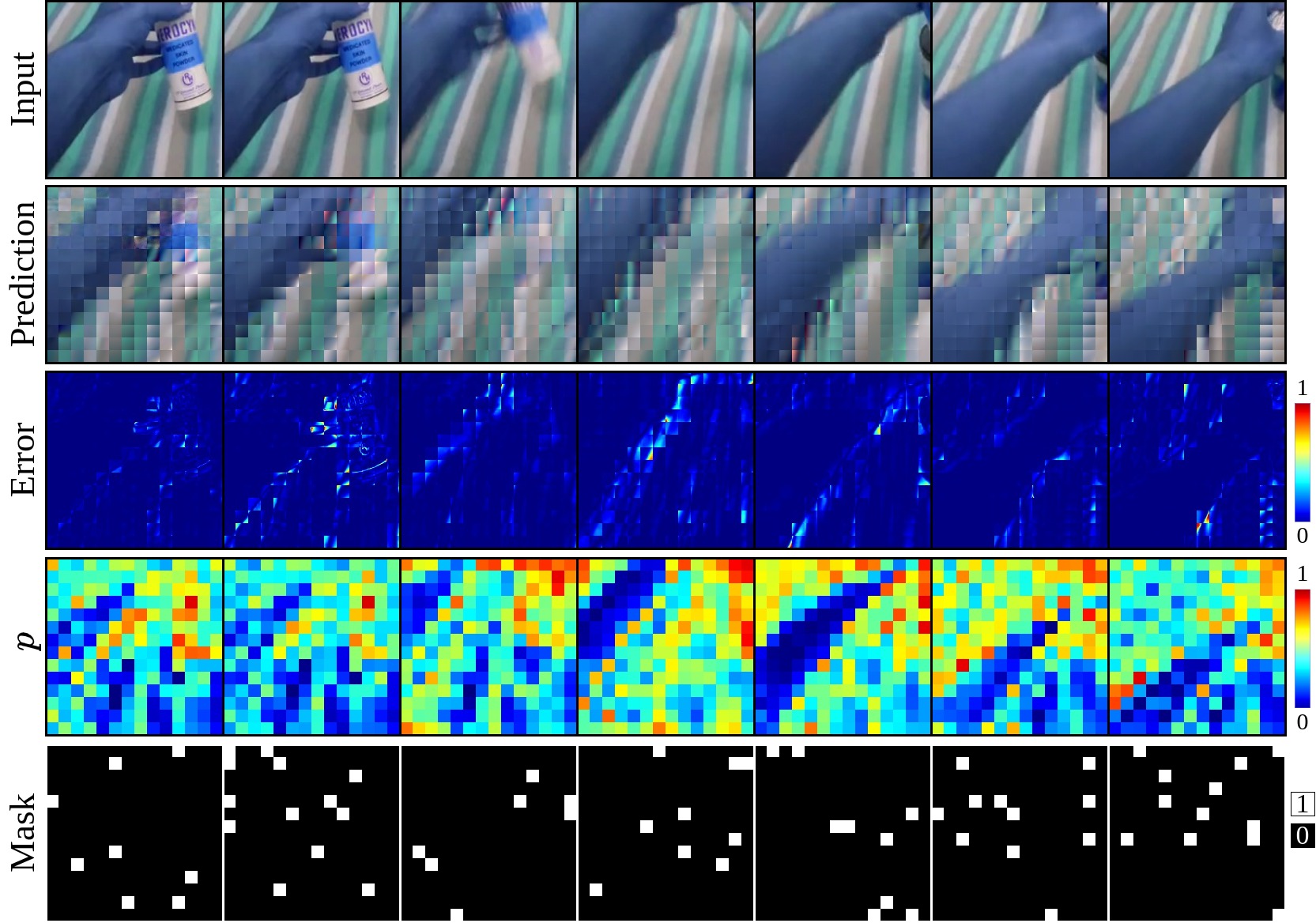}
    \caption{An example visualization of our adaptive sampling on SSv2 dataset.}
    \label{fig:ss_loss_4}
\end{figure}

\begin{figure}[tbh!]
    \centering
    \includegraphics[width=\linewidth]{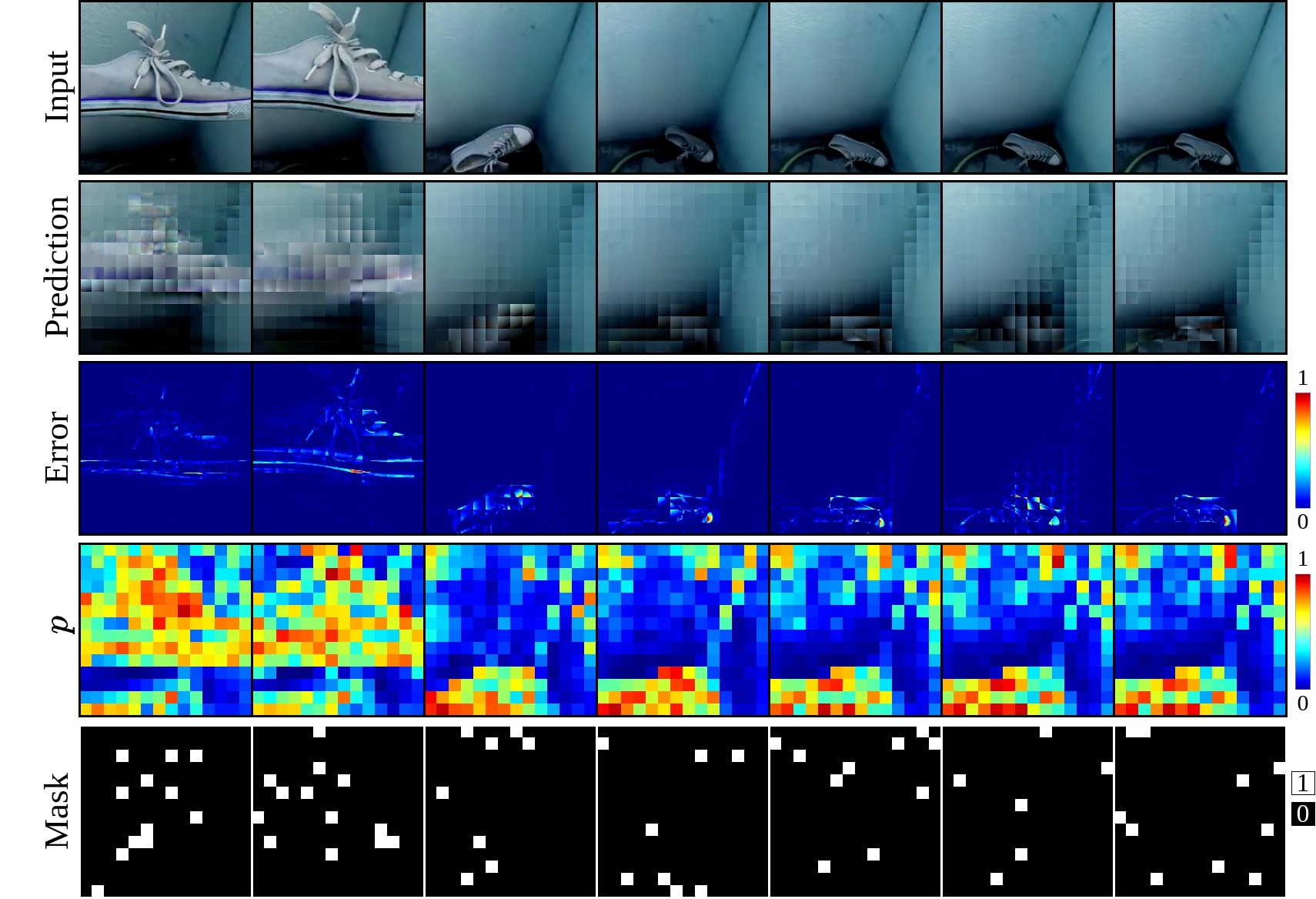}
    \caption{An example visualization of our adaptive sampling on SSv2 dataset.}
    \label{fig:ss_loss_5}
\end{figure}

\begin{figure}[tbh!]
    \centering
    \includegraphics[width=\linewidth]{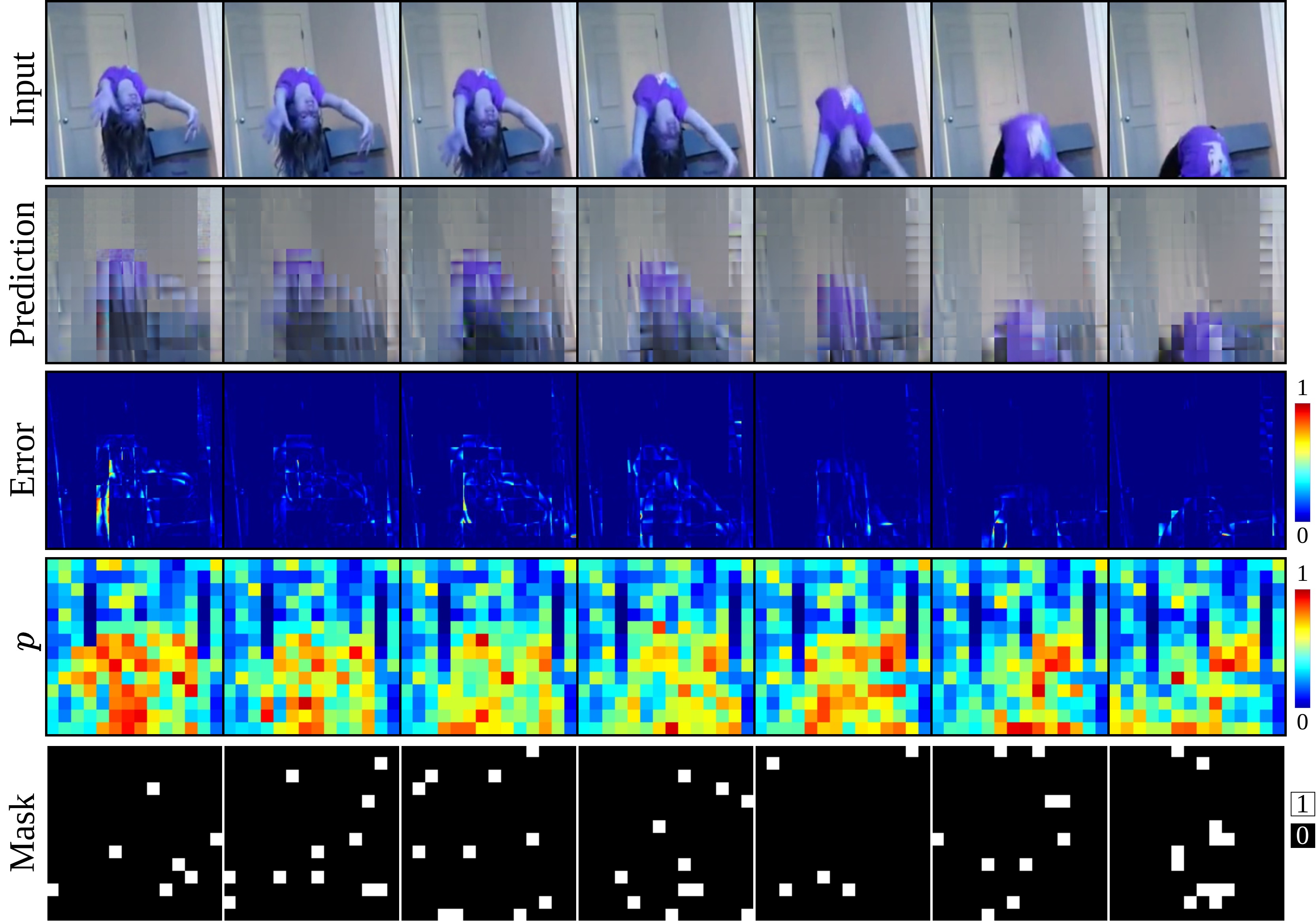}
    \caption{An example visualization of our adaptive sampling on K400 dataset.}
    \label{fig:k_loss_1}
\end{figure}

\begin{figure}[tbh!]
    \centering
    \includegraphics[width=\linewidth]{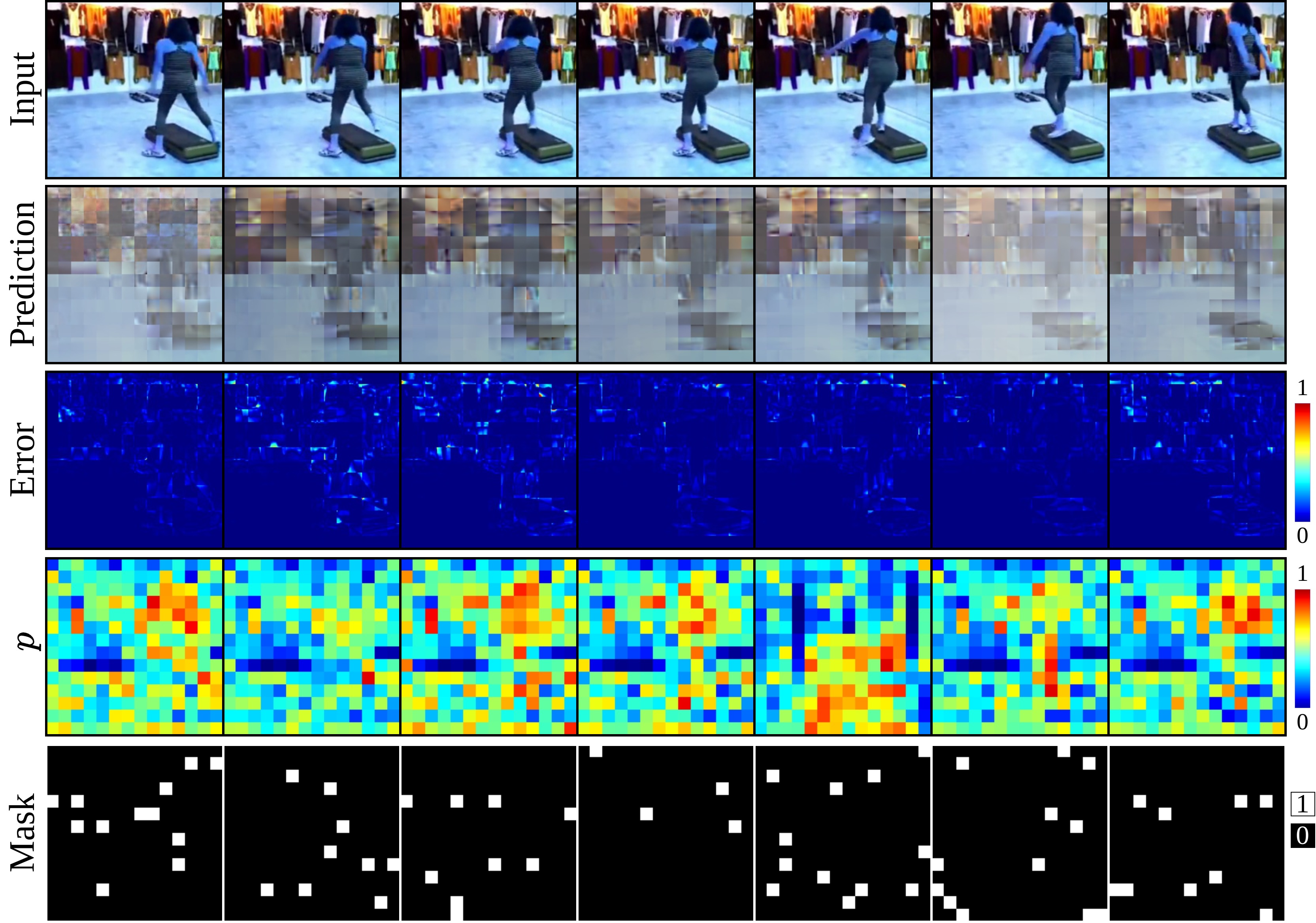}
    \caption{An example visualization of our adaptive sampling on K400 dataset.}
    \label{fig:k_loss_2}
\end{figure}

\begin{figure}[tbh!]
    \centering
    \includegraphics[width=\linewidth]{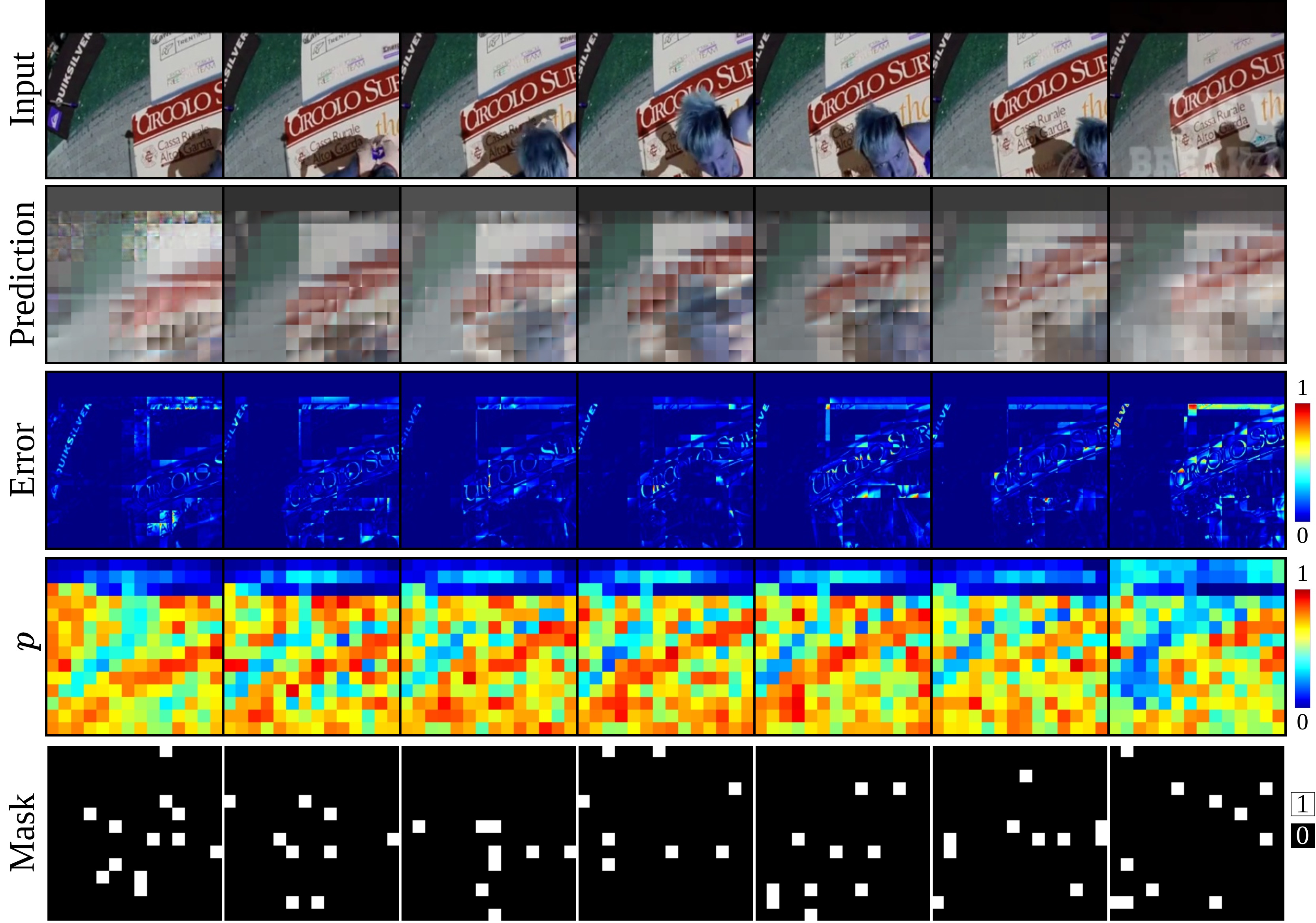}
    \caption{An example visualization of our adaptive sampling on K400 dataset.}
    \label{fig:k_loss_3}
\end{figure}

\begin{figure}[tbh!]
    \centering
    \includegraphics[width=\linewidth]{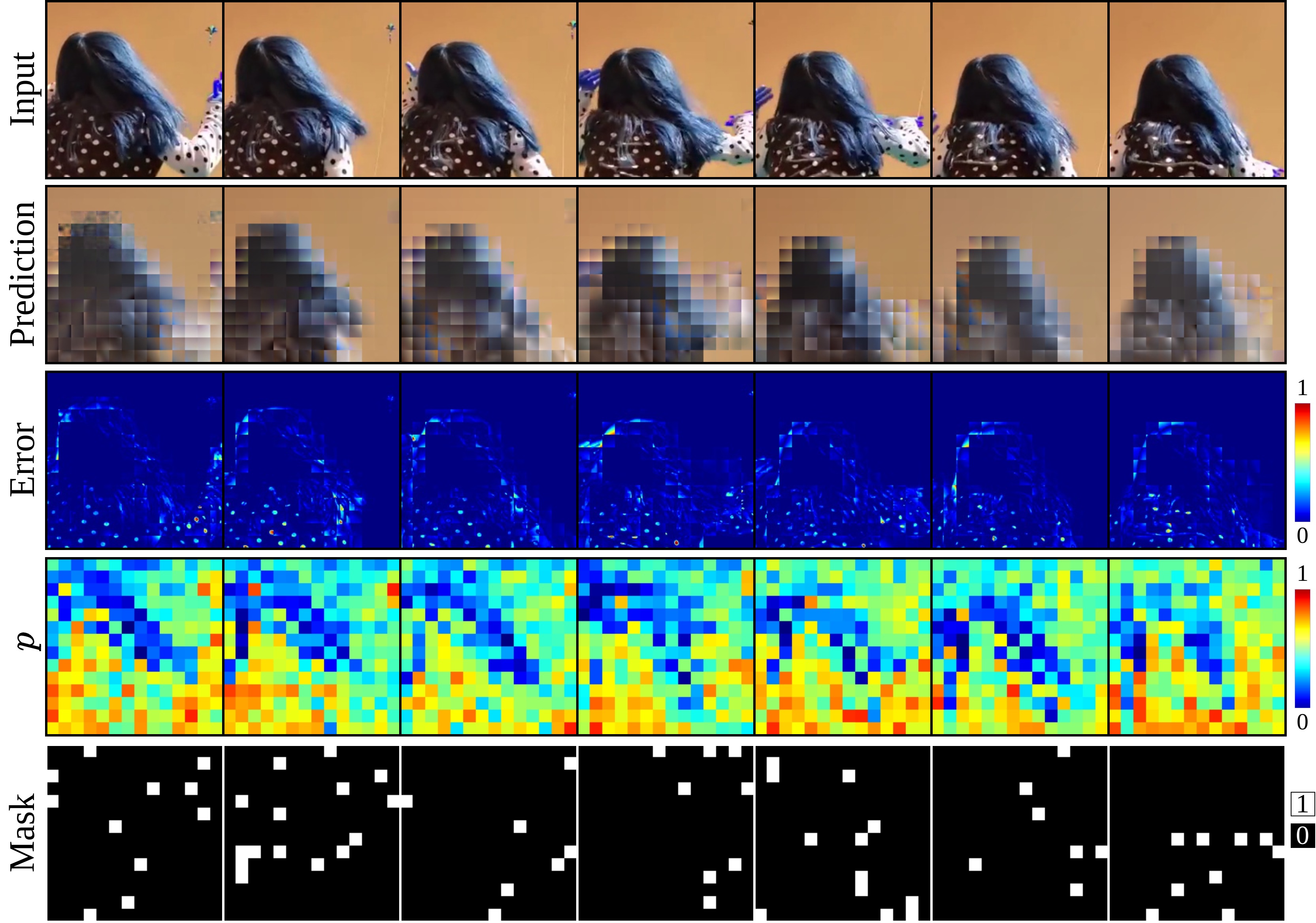}
    \caption{An example visualization of our adaptive sampling on K400 dataset.}
    \label{fig:k_loss_4}
\end{figure}

\begin{figure}[tbh!]
    \centering
    \includegraphics[width=\linewidth]{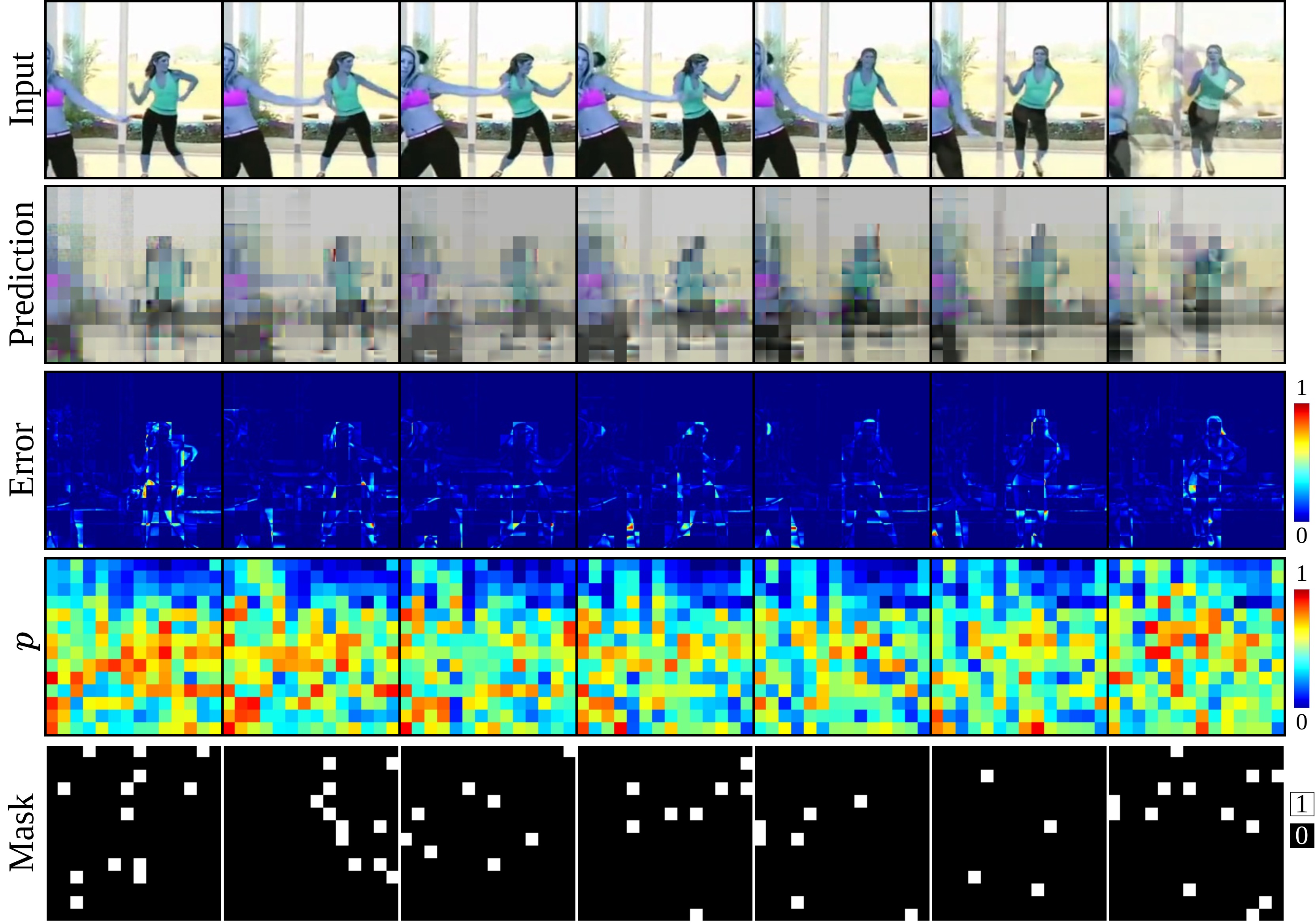}
    \caption{An example visualization of our adaptive sampling on K400 dataset.}
    \label{fig:k_loss_5}
\end{figure}

Here we provide some assitional visualizations to futher understand our AdaMAE for selected videos from SSv2 (Figure \ref{fig:ss_loss_1}, \ref{fig:ss_loss_2}, \ref{fig:ss_loss_3}, \ref{fig:ss_loss_4}, and \ref{fig:ss_loss_5}) and K400 (Figure \ref{fig:k_loss_1}, \ref{fig:k_loss_2}, \ref{fig:k_loss_3}, \ref{fig:k_loss_4}, and \ref{fig:k_loss_5}). 

For example, let us consider Figure \ref{fig:ss_loss_1}. We see that most of the spatiotemporal information is concentrated in the upper part of each frame (the region containing the bottle and the part of a hand), making those patches difficult to reconstruct accurately (see second and third rows). Since the proposed \texttt{AdaptiveSampling} network is optimized by maximizng $\mathbb{E}[L_R]$, it predicts higher probabilities for the patches from the high activity region. Hence, when sampling the visible tokens, MAE gets relatively more tokens from the high activity region than the low activity region as shown in the last row. 

We can make the similar observervations for the other examples as well.

\textit{Note that these results are for the default setting in Table \ref{tab:ablations} except here we visualize at 100-th epoch of pre-training.}

%% SSV2

% Kinetics

\clearpage
\subsection{Encoder-decoder architecture}
\label{sup:network_arch}
Table \ref{tab:encoder_decoder_arch} summarizes the encoder-decoder architecture of our AdaMAE. Specifically, we consider 16-frame valilla ViT-Base architecture for all experiments. We use assymetirc encoder-decoder architecture for self-supervised pre-training and discard the decoder during the fine-tuning.

Given a video we first extract 16 frames ($\textcolor{red}{3} \times \textcolor{blue}{16} \times \textcolor{green}{224} \times \textcolor{green}{224}$). We use temporal stride of $4$ and $2$ for K400 and SSv2 datasets, respectively.  Next we process these \textcolor{blue}{16} frames through a  Tokenizer which is essentially a 3D convolution layer with kernel size of $\textcolor{blue}{2} \times \textcolor{green}{16} \times \textcolor{green}{16}$, stride of $\textcolor{blue}{2} \times \textcolor{green}{16} \times \textcolor{green}{16}$ and output embedding dimention of \textcolor{red}{768}. This process results in a total of \textcolor{purple}{1568} tokens and each token is represented by a \textcolor{red}{768} dimention vector. Next, we sample $(1-\rho) \times \textcolor{purple}{1568}$ number of tokens as the visible tokens from our adaptive token sampling network. These visible tokens are then processed through the ViT encoder that comprises of 12 cascaded multi-head self-attention blocks (MHA blocks). The outputs from the ViT encoder is then combined with a fixed learnable representation for masked tokens which results in the \textcolor{purple}{1568} token represenrations. These \textcolor{purple}{1568} representations are then process through a projectot which brings down their embedding dimntion from \textcolor{red}{768} embedding dimention to \textcolor{red}{384} by an MLP layer. These projected representationas are then process through the ViT-decoder which consists of 4 MHA blocks followed by an MLP layer to bring the emebdding dimention from \textcolor{red}{384} to the total number of pixels in a cube which given by $\textcolor{blue}{2} \times \textcolor{red}{3} \times \textcolor{green}{16} \times \textcolor{green}{16} = \textcolor{green}{1536}$. This is finally reshaped back back to the original space and used to compute the reconsruction loss.

\begin{table}[tbh!]
    \centering
    \tablestyle{2.0pt}{1.04}
    \caption{Encoder-Decoder architecture of our AdaMAE. MHA denotes joint space-time multi-head self-attention.}
    \begin{tabular}{l|c|c}
    \shline\hline
    Stage & ViT-Base & Output shape \\
    \shline\hline
    \multirow{2}{*}{Input Video} & stride $\textcolor{blue}{4} \times \textcolor{green}{1} \times \textcolor{green}{1}$ for K400 & \multirow{2}{*}{$\textcolor{red}{3} \times \textcolor{blue}{16} \times \textcolor{green}{224} \times \textcolor{green}{224}$}\\
    & stride $\textcolor{blue}{2} \times \textcolor{green}{1} \times \textcolor{green}{1}$ for SSv2 & \\
    
    \hline
    \multirow{2}{*}{Tokenization} &  stride $\textcolor{blue}{2} \times \textcolor{green}{16} \times \textcolor{green}{16}$ & \multirow{2}{*}{$\textcolor{purple}{1568} \times \textcolor{red}{768} $}\\
    & emb. dim \textcolor{red}{768} & \\
    \hline
    
    \multirow{2}{*}{Masking} & Adaptive Masking & \multirow{2}{*}{$[(1-\rho) \times \textcolor{purple}{1568}] \times \textcolor{red}{768} $} \\
    & masking ratio $\rho$ & \\
    \hline
    
    \multirow{2}{*}{Encoder} & \multirow{2}{*}{$[MHA(\textcolor{red}{768})]\times 12$} &\multirow{2}{*}{$[(1-\rho) \times \textcolor{purple}{1568}] \times \textcolor{red}{768}$} \\ 
    & &\\
    \hline
    
    \multirow{2}{*}{Projection} & $MHA(384)$ &\multirow{2}{*}{$\textcolor{purple}{1568} \times \textcolor{red}{384}$} \\ 
    & concat masked tokens &\\
    \hline
    
    \multirow{2}{*}{Decoder} & \multirow{2}{*}{$[MHA(\textcolor{red}{384})]\times 4$} &\multirow{2}{*}{$[(1-\rho) \times \textcolor{purple}{1568}] \times \textcolor{red}{384}$} \\ 
    & &\\
    \hline
    
    \multirow{2}{*}{Projector} & \multirow{2}{*}{$MLP(\textcolor{red}{1536})$} &\multirow{2}{*}{$\textcolor{purple}{1568} \times \textcolor{red}{1536}$} \\ 
    &  &\\
    \hline
    
    \multirow{2}{*}{Reshaping} & \multirow{2}{*}{from 1536 to $\textcolor{red}{3} \times \textcolor{blue}{2} \times \textcolor{green}{16} \times \textcolor{green}{16}$} &\multirow{2}{*}{$\textcolor{red}{3} \times \textcolor{blue}{16} \times \textcolor{green}{224} \times \textcolor{green}{224}$} \\ 
    &  &\\
    \hline
    \shline\hline
    \end{tabular}
    \label{tab:encoder_decoder_arch}
\end{table}

\subsection{Pre-training setting}
\label{sup:pre_training}
Table \ref{tab:pretrain_ssv2_k400_config} summarizes the pre-training setting on SSv2 and K400 dataset.
In addition, we linearly scale the base learning rate with respect to the overall batch size, lr = base learning rate× batch size 256. We adopt the PyTorch and DeepSpeed frameworks for faster training.
\begin{table}[h]
    \centering
    \tablestyle{2.0pt}{1.04}
    \caption{Ptr-training setting on SSv2 and K400 datasets.}
    \begin{tabular}{l|cc}
    \shline\hline
    Configuration & SSv2 & K400 \\
    \shline\hline
    Optimizer & \multicolumn{2}{c}{Adamw}\\
    Optimizer betas & \multicolumn{2}{c}{\{0.9, 0.95\}}\\
    Base learning rate & \multicolumn{2}{c}{1.5e-4}\\
    Weight decay & \multicolumn{2}{c}{5e-2}\\
    Learning rate shedule & \multicolumn{2}{c}{cosine decay} \\
    Warmup epochs & \multicolumn{2}{c}{40}\\
    Flip augmentation & False & True \\
    Augmentation & \multicolumn{2}{c}{MultiScaleCrop}\\
    \shline\hline
    \end{tabular}
    \label{tab:pretrain_ssv2_k400_config}
\end{table}

\subsection{End-to-end fine tuning setting}
Table \ref{tab:finetune_ssv2_k400_config} summarizes the end-to-end fine-tunining setting on SSv2 and K400 datasets.
\begin{table}[h]
    \centering
    \tablestyle{2.0pt}{1.04}
    \caption{Ptr-training setting on SSv2 and K400 datasets.}
    \begin{tabular}{l|cc}
    \shline\hline
    Configuration & SSv2 & K400 \\
    \shline\hline
    Optimizer & \multicolumn{2}{c}{Adamw}\\
    Optimizer betas & \multicolumn{2}{c}{\{0.9, 0.999\}}\\
    Base learning rate & 1.5e-4 & 1e-3\\
    Layer-wse lr decay & 0.75 \\
    Weight decay &  \multicolumn{2}{c}{5e-2}\\
    Learning rate shedule & \multicolumn{2}{c}{cosine decay} \\
    Warmup epochs & \multicolumn{2}{c}{40}\\
    Flip augmentation & False & True \\
    RandAug & (0.9, 0.05)\\
    Label smoothing & 0.1 \\
    Mixup & 0.8 \\
    Cutmix & 1.0 \\
    drop path & 0.1 \\
    \shline\hline
    \end{tabular}
    \label{tab:finetune_ssv2_k400_config}
\end{table}

\subsection{Mask visualizations}
\label{sec:mask_vis_add}
\begin{figure*}[tbh!]
    \centering
    \includegraphics[width=0.8\linewidth]{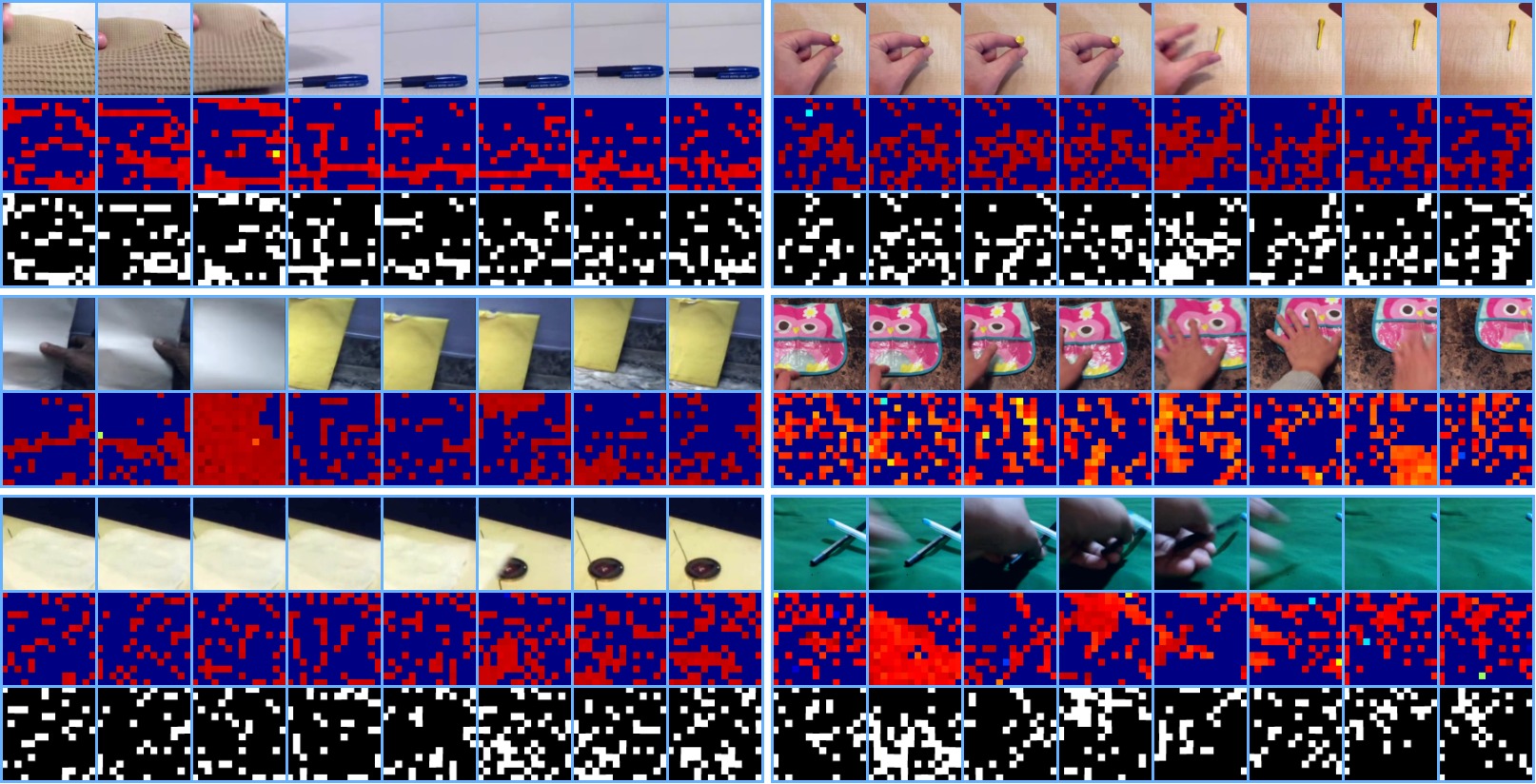}
    \caption{Mask visualizations of our AdaMAE for \textbf{80\% masking ratio} on SSV2 dataset~\cite{goyal_something_2017}.Given a video (\textit{first row}), our AdaMAE first predicts the categorical distribution (\textit{second row}), and then sample the mask (\textit{third row}) from that distribution. Colors closer \textcolor{red}{\bf red} and \textcolor{blue}{\bf blue} denotes the patches with \textit{high} and \textit{low} probability, respectively. In mask visualizations, \textbf{black} and \textbf{white} corresponds to the masked and visible path locations, respectively.}
    \label{fig:appendix_mask80}
\end{figure*}

\begin{figure*}[tbh!]
    \centering
    \includegraphics[width=0.8\linewidth]{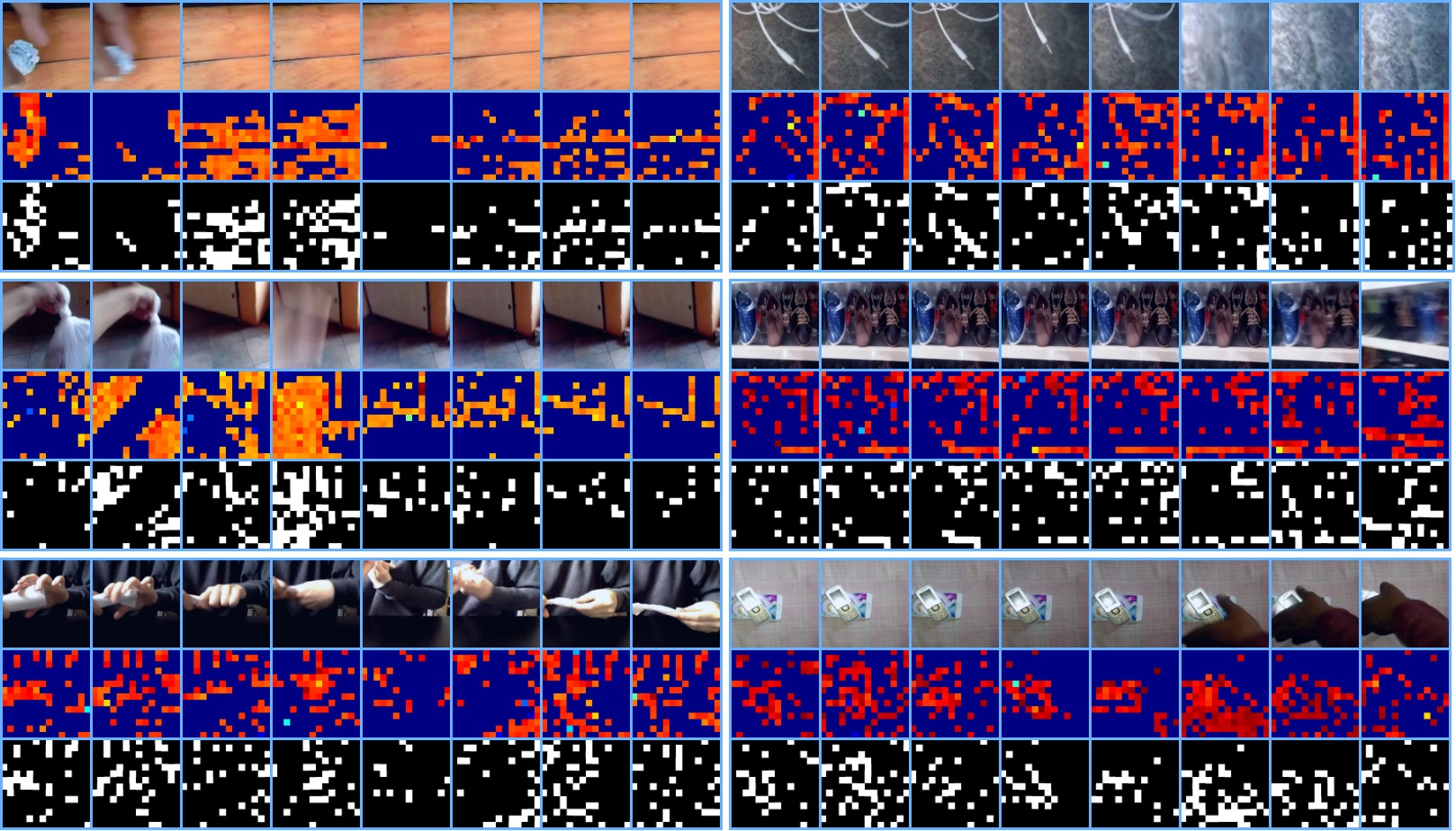}
    \caption{Mask visualizations of our AdaMAE for \textbf{85\% masking ratio} on SSV2 dataset~\cite{goyal_something_2017}.Given a video (\textit{first row}), our AdaMAE first predicts the categorical distribution (\textit{second row}), and then sample the mask (\textit{third row}) from that distribution. Colors closer \textcolor{red}{\bf red} and \textcolor{blue}{\bf blue} denotes the patches with \textit{high} and \textit{low} probability, respectively. In mask visualizations, \textbf{black} and \textbf{white} corresponds to the masked and visible path locations, respectively.}
    \label{fig:appendix_mask85}
\end{figure*}

\begin{figure*}[tbh!]
    \centering
    \includegraphics[width=0.8\linewidth]{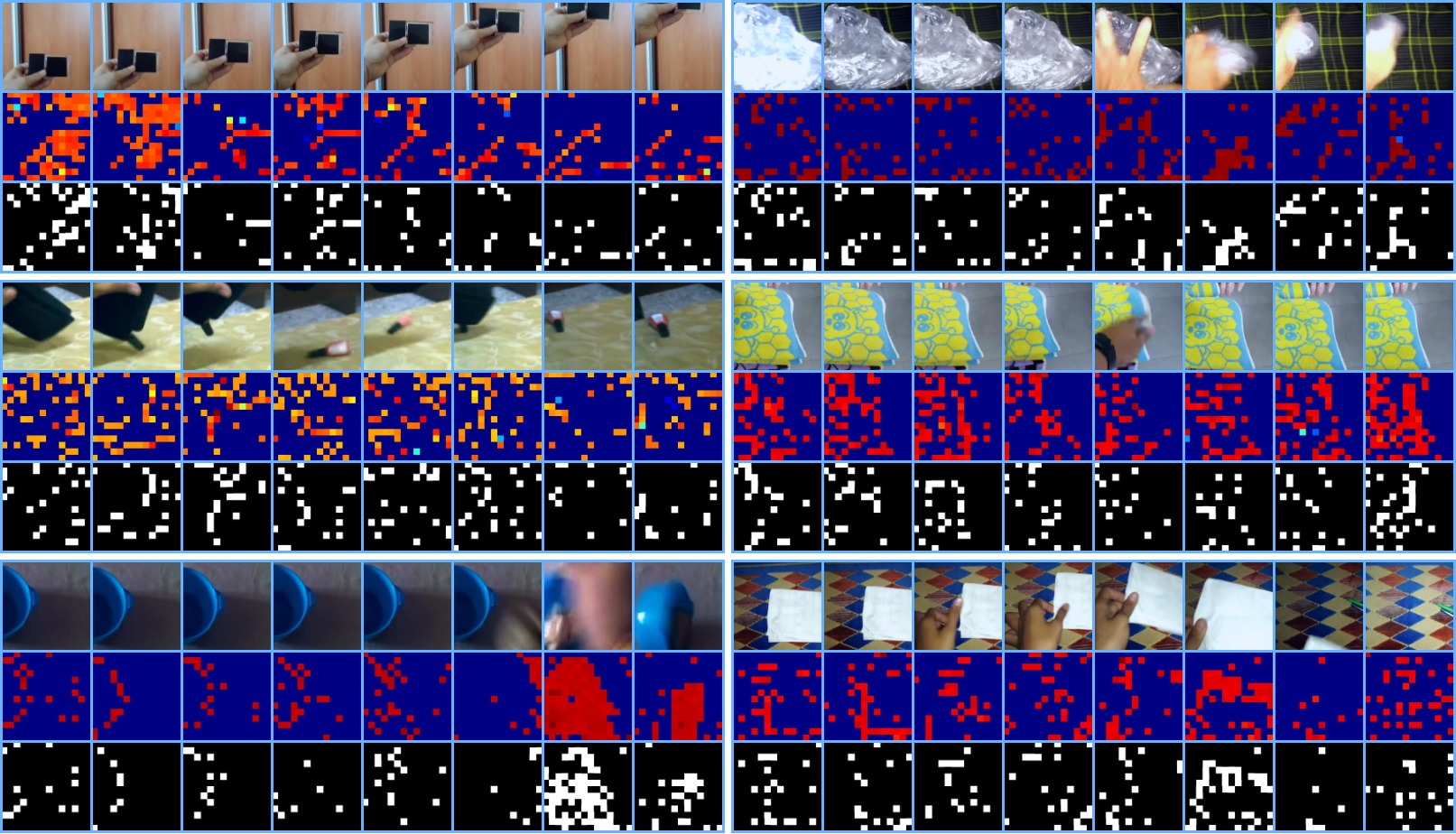}
    \caption{Mask visualizations of our AdaMAE for \textbf{90\% masking ratio} on SSV2 dataset~\cite{goyal_something_2017}.Given a video (\textit{first row}), our AdaMAE first predicts the categorical distribution (\textit{second row}), and then sample the mask (\textit{third row}) from that distribution. Colors closer \textcolor{red}{\bf red} and \textcolor{blue}{\bf blue} denotes the patches with \textit{high} and \textit{low} probability, respectively. In mask visualizations, \textbf{black} and \textbf{white} corresponds to the masked and visible path locations, respectively.}
    \label{fig:appendix_mask90}
\end{figure*}

\begin{figure*}[tbh!]
    \centering
    \includegraphics[width=0.8\linewidth]{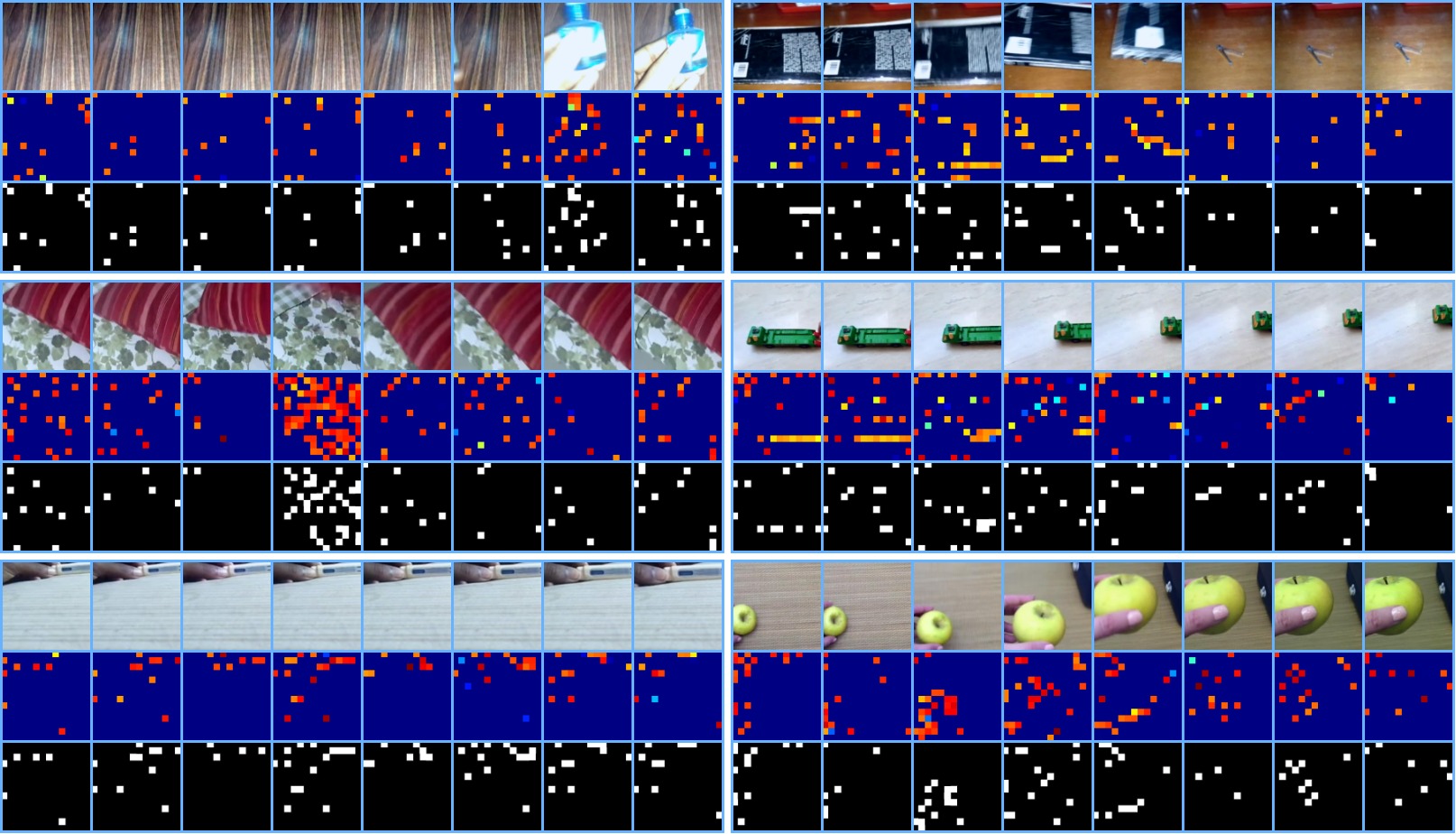}
    \caption{Mask visualizations of our AdaMAE for \textbf{95\% masking ratio} on SSV2 dataset~\cite{goyal_something_2017}.Given a video (\textit{first row}), our AdaMAE first predicts the categorical distribution (\textit{second row}), and then sample the mask (\textit{third row}) from that distribution. Colors closer \textcolor{red}{\bf red} and \textcolor{blue}{\bf blue} denotes the patches with \textit{high} and \textit{low} probability, respectively. In mask visualizations, \textbf{black} and \textbf{white} corresponds to the masked and visible path locations, respectively.}
    \label{fig:appendix_mask95}
\end{figure*}

Figure \ref{fig:appendix_mask80}, \ref{fig:appendix_mask85}, \ref{fig:appendix_mask90}, and \ref{fig:appendix_mask95} show the adaptive mask visualizations for masking ratio of 80\%, 85\%, 90\%, and 95\% for selected videos from SSv2 dataset.

\clearpage
% \subsection{Cross dataset experiments.}

% \subsection{Figures for MLP as mask sampling network}

%% file: 1-main.bbl
\begin{thebibliography}{10}\itemsep=-1pt

\bibitem{arnab_vivit_2021}
Anurag Arnab, Mostafa Dehghani, Georg Heigold, Chen Sun, Mario Lučić, and
  Cordelia Schmid.
\newblock {ViViT}: {A} {Video} {Vision} {Transformer}, Nov. 2021.
\newblock arXiv:2103.15691 [cs].

\bibitem{bao_beit_2022-1}
Hangbo Bao, Li Dong, Songhao Piao, and Furu Wei.
\newblock {BEiT}: {BERT} {Pre}-{Training} of {Image} {Transformers}, Sept.
  2022.
\newblock arXiv:2106.08254 [cs].

\bibitem{bao_beit_2022}
Hangbo Bao, Li Dong, Songhao Piao, and Furu Wei.
\newblock {BEiT}: {BERT} {Pre}-{Training} of {Image} {Transformers}, Sept.
  2022.
\newblock arXiv:2106.08254 [cs].

\bibitem{bertasius_is_2021}
Gedas Bertasius, Heng Wang, and Lorenzo Torresani.
\newblock Is {Space}-{Time} {Attention} {All} {You} {Need} for {Video}
  {Understanding}?, June 2021.
\newblock arXiv:2102.05095 [cs].

\bibitem{chen_efficient_2022-1}
Jun Chen, Ming Hu, Boyang Li, and Mohamed Elhoseiny.
\newblock Efficient {Self}-supervised {Vision} {Pretraining} with {Local}
  {Masked} {Reconstruction}, June 2022.
\newblock arXiv:2206.00790 [cs].

\bibitem{chen_generative_2020}
Mark Chen, Alec Radford, Rewon Child, Jeffrey Wu, Heewoo Jun, David Luan, and
  Ilya Sutskever.
\newblock Generative {Pretraining} {From} {Pixels}.
\newblock In Hal~Daumé III and Aarti Singh, editors, {\em Proceedings of the
  37th {International} {Conference} on {Machine} {Learning}}, volume 119 of
  {\em Proceedings of {Machine} {Learning} {Research}}, pages 1691--1703. PMLR,
  July 2020.

\bibitem{devlin_bert_2019}
Jacob Devlin, Ming-Wei Chang, Kenton Lee, and Kristina Toutanova.
\newblock {BERT}: {Pre}-training of {Deep} {Bidirectional} {Transformers} for
  {Language} {Understanding}, May 2019.
\newblock arXiv:1810.04805 [cs].

\bibitem{dosovitskiy_image_2021}
Alexey Dosovitskiy, Lucas Beyer, Alexander Kolesnikov, Dirk Weissenborn,
  Xiaohua Zhai, Thomas Unterthiner, Mostafa Dehghani, Matthias Minderer, Georg
  Heigold, Sylvain Gelly, Jakob Uszkoreit, and Neil Houlsby.
\newblock An {Image} is {Worth} 16x16 {Words}: {Transformers} for {Image}
  {Recognition} at {Scale}, June 2021.
\newblock arXiv:2010.11929 [cs].

\bibitem{fan_multiscale_2021}
Haoqi Fan, Bo Xiong, Karttikeya Mangalam, Yanghao Li, Zhicheng Yan, Jitendra
  Malik, and Christoph Feichtenhofer.
\newblock Multiscale {Vision} {Transformers}, Apr. 2021.
\newblock arXiv:2104.11227 [cs].

\bibitem{fang_unleashing_2022}
Yuxin Fang, Shusheng Yang, Shijie Wang, Yixiao Ge, Ying Shan, and Xinggang
  Wang.
\newblock Unleashing {Vanilla} {Vision} {Transformer} with {Masked} {Image}
  {Modeling} for {Object} {Detection}, May 2022.
\newblock arXiv:2204.02964 [cs].

\bibitem{feichtenhofer_masked_2022}
Christoph Feichtenhofer, Haoqi Fan, Yanghao Li, and Kaiming He.
\newblock Masked {Autoencoders} {As} {Spatiotemporal} {Learners}, May 2022.
\newblock arXiv:2205.09113 [cs].

\bibitem{feichtenhofer_slowfast_2019}
Christoph Feichtenhofer, Haoqi Fan, Jitendra Malik, and Kaiming He.
\newblock {SlowFast} {Networks} for {Video} {Recognition}, Oct. 2019.
\newblock arXiv:1812.03982 [cs].

\bibitem{gao_convmae_2022}
Peng Gao, Teli Ma, Hongsheng Li, Ziyi Lin, Jifeng Dai, and Yu Qiao.
\newblock {ConvMAE}: {Masked} {Convolution} {Meets} {Masked} {Autoencoders},
  May 2022.
\newblock arXiv:2205.03892 [cs].

\bibitem{girdhar_omnimae_2022}
Rohit Girdhar, Alaaeldin El-Nouby, Mannat Singh, Kalyan~Vasudev Alwala, Armand
  Joulin, and Ishan Misra.
\newblock {OmniMAE}: {Single} {Model} {Masked} {Pretraining} on {Images} and
  {Videos}, June 2022.
\newblock arXiv:2206.08356 [cs, stat].

\bibitem{goyal_something_2017}
Raghav Goyal, Samira~Ebrahimi Kahou, Vincent Michalski, Joanna Materzyńska,
  Susanne Westphal, Heuna Kim, Valentin Haenel, Ingo Fruend, Peter Yianilos,
  Moritz Mueller-Freitag, Florian Hoppe, Christian Thurau, Ingo Bax, and Roland
  Memisevic.
\newblock The "something something" video database for learning and evaluating
  visual common sense, June 2017.
\newblock arXiv:1706.04261 [cs].

\bibitem{gupta_maskvit_2022}
Agrim Gupta, Stephen Tian, Yunzhi Zhang, Jiajun Wu, Roberto Martín-Martín,
  and Li Fei-Fei.
\newblock {MaskViT}: {Masked} {Visual} {Pre}-{Training} for {Video}
  {Prediction}, Aug. 2022.
\newblock arXiv:2206.11894 [cs].

\bibitem{he_masked_2021}
Kaiming He, Xinlei Chen, Saining Xie, Yanghao Li, Piotr Dollár, and Ross
  Girshick.
\newblock Masked {Autoencoders} {Are} {Scalable} {Vision} {Learners}, Dec.
  2021.
\newblock arXiv:2111.06377 [cs].

\bibitem{hendrycks_using_2019}
Dan Hendrycks, Mantas Mazeika, Saurav Kadavath, and Dawn Song.
\newblock Using {Self}-{Supervised} {Learning} {Can} {Improve} {Model}
  {Robustness} and {Uncertainty}.
\newblock In {\em Advances in {Neural} {Information} {Processing} {Systems}},
  volume~32. Curran Associates, Inc., 2019.

\bibitem{hou_milan_2022}
Zejiang Hou, Fei Sun, Yen-Kuang Chen, Yuan Xie, and Sun-Yuan Kung.
\newblock {MILAN}: {Masked} {Image} {Pretraining} on {Language} {Assisted}
  {Representation}, Aug. 2022.
\newblock arXiv:2208.06049 [cs].

\bibitem{huang_green_2022}
Lang Huang, Shan You, Mingkai Zheng, Fei Wang, Chen Qian, and Toshihiko
  Yamasaki.
\newblock Green {Hierarchical} {Vision} {Transformer} for {Masked} {Image}
  {Modeling}, May 2022.
\newblock arXiv:2205.13515 [cs].

\bibitem{jaiswal_survey_2021}
Ashish Jaiswal, Ashwin~Ramesh Babu, Mohammad~Zaki Zadeh, Debapriya Banerjee,
  and Fillia Makedon.
\newblock A {Survey} on {Contrastive} {Self}-{Supervised} {Learning}.
\newblock {\em Technologies}, 9(1):2, Mar. 2021.
\newblock Number: 1 Publisher: Multidisciplinary Digital Publishing Institute.

\bibitem{jing_understanding_2022}
Li Jing, Pascal Vincent, Yann LeCun, and Yuandong Tian.
\newblock Understanding {Dimensional} {Collapse} in {Contrastive}
  {Self}-supervised {Learning}, Apr. 2022.
\newblock arXiv:2110.09348 [cs].

\bibitem{kaelbling_reinforcement_1996}
L.~P. Kaelbling, M.~L. Littman, and A.~W. Moore.
\newblock Reinforcement {Learning}: {A} {Survey}.
\newblock {\em Journal of Artificial Intelligence Research}, 4:237--285, May
  1996.

\bibitem{kay_kinetics_2017}
Will Kay, Joao Carreira, Karen Simonyan, Brian Zhang, Chloe Hillier, Sudheendra
  Vijayanarasimhan, Fabio Viola, Tim Green, Trevor Back, Paul Natsev, Mustafa
  Suleyman, and Andrew Zisserman.
\newblock The {Kinetics} {Human} {Action} {Video} {Dataset}, May 2017.
\newblock arXiv:1705.06950 [cs].

\bibitem{lan_deep_2018}
Xu Lan, Hanxiao Wang, Shaogang Gong, and Xiatian Zhu.
\newblock Deep {Reinforcement} {Learning} {Attention} {Selection} for {Person}
  {Re}-{Identification}, July 2018.
\newblock arXiv:1707.02785 [cs].

\bibitem{lee_stochastic_2020}
Alex~X. Lee, Anusha Nagabandi, Pieter Abbeel, and Sergey Levine.
\newblock Stochastic {Latent} {Actor}-{Critic}: {Deep} {Reinforcement}
  {Learning} with a {Latent} {Variable} {Model}, Oct. 2020.
\newblock arXiv:1907.00953 [cs, stat].

\bibitem{lin_tsm_2019}
Ji Lin, Chuang Gan, and Song Han.
\newblock {TSM}: {Temporal} {Shift} {Module} for {Efficient} {Video}
  {Understanding}, Aug. 2019.
\newblock arXiv:1811.08383 [cs].

\bibitem{liu_teinet_2019}
Zhaoyang Liu, Donghao Luo, Yabiao Wang, Limin Wang, Ying Tai, Chengjie Wang,
  Jilin Li, Feiyue Huang, and Tong Lu.
\newblock {TEINet}: {Towards} an {Efficient} {Architecture} for {Video}
  {Recognition}, Nov. 2019.
\newblock arXiv:1911.09435 [cs].

\bibitem{liu_video_2021}
Ze Liu, Jia Ning, Yue Cao, Yixuan Wei, Zheng Zhang, Stephen Lin, and Han Hu.
\newblock Video {Swin} {Transformer}, June 2021.
\newblock arXiv:2106.13230 [cs].

\bibitem{liu_tam_2021}
Zhaoyang Liu, Limin Wang, Wayne Wu, Chen Qian, and Tong Lu.
\newblock {TAM}: {Temporal} {Adaptive} {Module} for {Video} {Recognition}, Aug.
  2021.
\newblock arXiv:2005.06803 [cs].

\bibitem{loshchilov_decoupled_2019}
Ilya Loshchilov and Frank Hutter.
\newblock Decoupled {Weight} {Decay} {Regularization}, Jan. 2019.
\newblock arXiv:1711.05101 [cs, math].

\bibitem{marriott__charles_dictionary_1990}
Francis~Henry Marriott, Charles.
\newblock {\em A dictionary of statistical terms.}
\newblock Longman Scientific \& Technical, 5-th edition, 1990.

\bibitem{mnih_playing_2013}
Volodymyr Mnih, Koray Kavukcuoglu, David Silver, Alex Graves, Ioannis
  Antonoglou, Daan Wierstra, and Martin Riedmiller.
\newblock Playing {Atari} with {Deep} {Reinforcement} {Learning}, Dec. 2013.
\newblock arXiv:1312.5602 [cs].

\bibitem{neimark_video_2021}
Daniel Neimark, Omri Bar, Maya Zohar, and Dotan Asselmann.
\newblock Video {Transformer} {Network}, Aug. 2021.
\newblock arXiv:2102.00719 [cs].

\bibitem{oord_representation_2019}
Aaron van~den Oord, Yazhe Li, and Oriol Vinyals.
\newblock Representation {Learning} with {Contrastive} {Predictive} {Coding},
  Jan. 2019.
\newblock arXiv:1807.03748 [cs, stat].

\bibitem{patrick_keeping_2021}
Mandela Patrick, Dylan Campbell, Yuki~M. Asano, Ishan Misra, Florian Metze,
  Christoph Feichtenhofer, Andrea Vedaldi, and João~F. Henriques.
\newblock Keeping {Your} {Eye} on the {Ball}: {Trajectory} {Attention} in
  {Video} {Transformers}, Oct. 2021.
\newblock arXiv:2106.05392 [cs].

\bibitem{qian_spatiotemporal_2021}
Rui Qian, Tianjian Meng, Boqing Gong, Ming-Hsuan Yang, Huisheng Wang, Serge
  Belongie, and Yin Cui.
\newblock Spatiotemporal {Contrastive} {Video} {Representation} {Learning}.
\newblock In {\em 2021 {IEEE}/{CVF} {Conference} on {Computer} {Vision} and
  {Pattern} {Recognition} ({CVPR})}, pages 6960--6970, Nashville, TN, USA, June
  2021. IEEE.

\bibitem{radford_improving_nodate}
Alec Radford, Karthik Narasimhan, Tim Salimans, and Ilya Sutskever.
\newblock Improving {Language} {Understanding} by {Generative}
  {Pre}-{Training}.
\newblock {\em cs.ubc}, page~12, 2018.

\bibitem{ramesh_zero-shot_2021}
Aditya Ramesh, Mikhail Pavlov, Gabriel Goh, Scott Gray, Chelsea Voss, Alec
  Radford, Mark Chen, and Ilya Sutskever.
\newblock Zero-{Shot} {Text}-to-{Image} {Generation}, Feb. 2021.
\newblock arXiv:2102.12092 [cs].

\bibitem{shelhamer_loss_2017}
Evan Shelhamer, Parsa Mahmoudieh, Max Argus, and Trevor Darrell.
\newblock Loss is its own {Reward}: {Self}-{Supervision} for {Reinforcement}
  {Learning}, Mar. 2017.
\newblock arXiv:1612.07307 [cs].

\bibitem{song_category_2016}
Xinhui Song, Ke Chen, Jie Lei, Li Sun, Zhiyuan Wang, Lei Xie, and Mingli Song.
\newblock Category driven deep recurrent neural network for video
  summarization.
\newblock In {\em 2016 {IEEE} {International} {Conference} on {Multimedia} \&
  {Expo} {Workshops} ({ICMEW})}, pages 1--6, July 2016.

\bibitem{sutton_policy_1999}
Richard~S Sutton, David McAllester, Satinder Singh, and Yishay Mansour.
\newblock Policy {Gradient} {Methods} for {Reinforcement} {Learning} with
  {Function} {Approximation}.
\newblock In {\em Advances in {Neural} {Information} {Processing} {Systems}},
  volume~12. MIT Press, 1999.

\bibitem{szegedy_going_2014}
Christian Szegedy, Wei Liu, Yangqing Jia, Pierre Sermanet, Scott Reed, Dragomir
  Anguelov, Dumitru Erhan, Vincent Vanhoucke, and Andrew Rabinovich.
\newblock Going {Deeper} with {Convolutions}, Sept. 2014.
\newblock arXiv:1409.4842 [cs].

\bibitem{tan_vimpac_2021}
Hao Tan, Jie Lei, Thomas Wolf, and Mohit Bansal.
\newblock {VIMPAC}: {Video} {Pre}-{Training} via {Masked} {Token} {Prediction}
  and {Contrastive} {Learning}, June 2021.
\newblock arXiv:2106.11250 [cs].

\bibitem{tao_siamese_2022}
Chenxin Tao, Xizhou Zhu, Gao Huang, Yu Qiao, Xiaogang Wang, and Jifeng Dai.
\newblock Siamese {Image} {Modeling} for {Self}-{Supervised} {Vision}
  {Representation} {Learning}, July 2022.
\newblock arXiv:2206.01204 [cs].

\bibitem{tong_videomae_2022}
Zhan Tong, Yibing Song, Jue Wang, and Limin Wang.
\newblock {VideoMAE}: {Masked} {Autoencoders} are {Data}-{Efficient} {Learners}
  for {Self}-{Supervised} {Video} {Pre}-{Training}, July 2022.
\newblock arXiv:2203.12602 [cs].

\bibitem{vaswani_attention_2017}
Ashish Vaswani, Noam Shazeer, Niki Parmar, Jakob Uszkoreit, Llion Jones,
  Aidan~N. Gomez, Lukasz Kaiser, and Illia Polosukhin.
\newblock Attention {Is} {All} {You} {Need}, Dec. 2017.
\newblock arXiv:1706.03762 [cs].

\bibitem{wang_tdn_2021}
Limin Wang, Zhan Tong, Bin Ji, and Gangshan Wu.
\newblock {TDN}: {Temporal} {Difference} {Networks} for {Efficient} {Action}
  {Recognition}, Mar. 2021.
\newblock arXiv:2012.10071 [cs].

\bibitem{wang_bevt_2022}
Rui Wang, Dongdong Chen, Zuxuan Wu, Yinpeng Chen, Xiyang Dai, Mengchen Liu,
  Yu-Gang Jiang, Luowei Zhou, and Lu Yuan.
\newblock {BEVT}: {BERT} {Pretraining} of {Video} {Transformers}, Mar. 2022.
\newblock arXiv:2112.01529 [cs].

\bibitem{wang_non-local_2018}
Xiaolong Wang, Ross Girshick, Abhinav Gupta, and Kaiming He.
\newblock Non-local {Neural} {Networks}, Apr. 2018.
\newblock arXiv:1711.07971 [cs].

\bibitem{wei_masked_2021}
Chen Wei, Haoqi Fan, Saining Xie, Chao-Yuan Wu, Alan Yuille, and Christoph
  Feichtenhofer.
\newblock Masked {Feature} {Prediction} for {Self}-{Supervised} {Visual}
  {Pre}-{Training}, Dec. 2021.
\newblock arXiv:2112.09133 [cs].

\bibitem{wettig_should_2022}
Alexander Wettig, Tianyu Gao, Zexuan Zhong, and Danqi Chen.
\newblock Should {You} {Mask} 15\% in {Masked} {Language} {Modeling}?, May
  2022.
\newblock arXiv:2202.08005 [cs].

\bibitem{williams_simple_1992}
Ronald~J. Williams.
\newblock Simple statistical gradient-following algorithms for connectionist
  reinforcement learning.
\newblock {\em Machine Learning}, 8(3):229--256, May 1992.

\bibitem{xie_masked_2022}
Jiahao Xie, Wei Li, Xiaohang Zhan, Ziwei Liu, Yew~Soon Ong, and Chen~Change
  Loy.
\newblock Masked {Frequency} {Modeling} for {Self}-{Supervised} {Visual}
  {Pre}-{Training}, June 2022.
\newblock arXiv:2206.07706 [cs].

\bibitem{xie_simmim_2022}
Zhenda Xie, Zheng Zhang, Yue Cao, Yutong Lin, Jianmin Bao, Zhuliang Yao, Qi
  Dai, and Han Hu.
\newblock {SimMIM}: {A} {Simple} {Framework} for {Masked} {Image} {Modeling},
  Apr. 2022.
\newblock arXiv:2111.09886 [cs].

\bibitem{xu_show_2016}
Kelvin Xu, Jimmy Ba, Ryan Kiros, Kyunghyun Cho, Aaron Courville, Ruslan
  Salakhutdinov, Richard Zemel, and Yoshua Bengio.
\newblock Show, {Attend} and {Tell}: {Neural} {Image} {Caption} {Generation}
  with {Visual} {Attention}, Apr. 2016.
\newblock arXiv:1502.03044 [cs].

\bibitem{yan_multiview_2022}
Shen Yan, Xuehan Xiong, Anurag Arnab, Zhichao Lu, Mi Zhang, Chen Sun, and
  Cordelia Schmid.
\newblock Multiview {Transformers} for {Video} {Recognition}, May 2022.
\newblock arXiv:2201.04288 [cs].

\bibitem{zhang_survey_2022}
Chaoning Zhang, Chenshuang Zhang, Junha Song, John Seon~Keun Yi, Kang Zhang,
  and In~So Kweon.
\newblock A {Survey} on {Masked} {Autoencoder} for {Self}-supervised {Learning}
  in {Vision} and {Beyond}, July 2022.
\newblock arXiv:2208.00173 [cs].

\bibitem{zhang_dual_2022}
Chaoning Zhang, Kang Zhang, Trung~X. Pham, Axi Niu, Zhinan Qiao, Chang~D. Yoo,
  and In~So Kweon.
\newblock Dual {Temperature} {Helps} {Contrastive} {Learning} {Without} {Many}
  {Negative} {Samples}: {Towards} {Understanding} and {Simplifying} {MoCo},
  Mar. 2022.
\newblock arXiv:2203.17248 [cs].

\bibitem{zhang_how_2022}
Chaoning Zhang, Kang Zhang, Chenshuang Zhang, Trung~X. Pham, Chang~D. Yoo, and
  In~So Kweon.
\newblock How {Does} {SimSiam} {Avoid} {Collapse} {Without} {Negative}
  {Samples}? {A} {Unified} {Understanding} with {Self}-supervised {Contrastive}
  {Learning}, Mar. 2022.
\newblock arXiv:2203.16262 [cs].

\bibitem{zhou_deep_2018}
Kaiyang Zhou, Yu Qiao, and Tao Xiang.
\newblock Deep {Reinforcement} {Learning} for {Unsupervised} {Video}
  {Summarization} with {Diversity}-{Representativeness} {Reward}, Feb. 2018.
\newblock arXiv:1801.00054 [cs] version: 3.

\end{thebibliography}
